# ACADEMIC WRITING WITH GPT-3.5: REFLECTIONS ON PRACTICES, EFFICACY AND TRANSPARENCY

**Oğuz 'Oz' Buruk**
Tampere University
Tampere, Finland
oguz.buruk@tuni.fi

**ABSTRACT**
The debate around the use of GPT-3.5 has been a popular topic among academics since the release of ChatGPT. Whilst some have argued for the advantages of GPT-3.5 in enhancing academic writing, others have raised concerns such as plagiarism, the spread of false information, and ecological issues. The need for finding ways to use GPT-3.5 models transparently has been voiced, and suggestions have been made on social media as to how to use GPT-3.5 models in a smart way. Nevertheless, to date, there is a lack of literature which clearly outlines how to use GPT-3.5 models in academic writing, how effective they are, and how to use them transparently. To address this, I conducted a personal experience experiment with GPT-3.5, specifically by using text-davinci-003 model of OpenAI, for writing this article. I identified five ways of using GPT-3.5: Chunk Stylist, Bullet-to-Paragraph, Talk Textualizer, Research Buddy, and Polisher. I reflected on their efficacy, and commented on their potential impact on writing ethics. Additionally, I provided a comprehensive document which shows the prompts I used, results I got from GPT-3.5, the final edits and visually compares those by showing the differences in percentage.

**INTRODUCTION**
In recent months, GPT-3.5 models has become very popular among writers and researchers, especially with the lunch of ChatGPT. There has been a surge of interest in using this tool for academic writing for a variety of purposes. Scientists have even gone as far as to add ChatGPT as a co-author to papers.

The ability to communicate scientific material in a paper is essential for successful writing. In some cases, the lack of skill, hesitation, or being slow in or disliking writing process can present significant barriers for researchers independent from their experience [14]. The potential of GPT-3.5 therefore to revolutionize the way science is communicated cannot be underestimated. In addition, it offers a way for those with limited writing skills, or whose mother tongue is not English, to flowingly put their ideas into words, which can help to reduce the advantage gap between researchers who do and do not have English as their first language [8].

Notwithstanding the excitement surrounding GPT-3.5, it has also raised serious concerns among scientists and journals. Notable outlets such as Nature and Science have published statements expressing the unease with its utilization [17,22], as well as the main principles for its use. Institutions have also published guidelines for employing GPT-3.5 in research and education [13]. These are mainly focused on ethical issues, such as content ownership, plagiarism and the potential for misinformation due to careless use without a proper check of a human-author.

In order for GPT-3.5 to be employed in an efficient yet responsible manner, further studies must be conducted to demonstrate how it can best be used. This piece details a personal experience experiment where I used GPT-3.5 in various ways in the production of this article. It reflects on the efficacy of the tool, outlines different ways of incorporating it into the writing process, and discusses methods for ethical and responsible use. I did not use ChatGPT, and instead used the Open AI Playground Interface which provides more control such as Temperature settings for changing the randomness level of the outcome. Both ChatGPT and Playground are based on the same training models and I specifically used text-davinci-003 in this experiment.

Through this experiment, I hope to provide researchers with:

- A better understanding of how GPT-3.5 can facilitate effective academic writing
- Possible ways to use it
- Reflections on ethical use and transparency

**GPT-3 IN ACADEMIC WRITING**
GPT-3.5 is a natural language processing (NLP) model developed by OpenAI, based on a deep learning technique called transformers. It is a large-scale language model that is trained on a massive amount of text data, allowing it to generate human-like text. GPT-3.5 is capable of generating text that is coherent and consistent with the context it is provided. It can be used for a variety of tasks, including text summarization, question answering, and text generation. Currently, code-davinci-002, text-davinci-002 and text-davinci-003 (which is also used in the generation of this paper) models are referred to as GPT 3.5. The popular ChatGPT application, and other services of OpenAI Payground such as text completion, incorporates models that can be reffered to GPT-3.5 [23].

GPT-3.5, and its predecessor like GPT-3 and GPT-2, have recently become popular amongst academics for their utilization in different scientific communication purposes such as addressing challenges in writing, navigating extensive literature and providing definitions of concepts [12]. Twitter threads have been released demonstrating the potential uses of ChatGPT in a "smart" way [1]. Studies conducted by independent experts to assess the quality of content found that it produces high-quality results, which are hard to distinguish from human-generated content [19]. GPT models has also been proposed as an aid for academic writing for students [15].

Despite the potential uses of GPT in academic writing, there are other potentially problematic results that have been surfaced. For example, some researchers have added ChatGPT as a co-author in research papers [9,10]. Such actions have been advised against by Science and Nature, who have released editorials [17,22] to dispel the notion of GPT authorship, citing concerns such as assigning responsibility, agency and ownership of content produced by language learning models, plagiarism and transparency. Debouche have also raised similar concerns for the utilization of GPT-3 and recommended authors to openly share the prompts and outcomes used [5]. Other issues include embedded biases (such as hate speech towards race, sexism) [3,11], exploitation of workers for data labeling, permissions on the data used for training and environmental concerns due to the energy used in algorithm training [16]. Moreover, some studies have proposed that AI-supported NLP models should be open-source and developed together with stakeholders, rather than being a product of a private company [4].

In conclusion, although GPT models has been proposed as a tool for academic writing, there are still debates focusing on the potential positives and negatives of its utilization. We also lack studies that would reflect on its efficacy in supporting writing, different ways of utilizing them and demonstrate a workflow that can be considered transparent by reviewers.

**PROCEDURE FOLLOWED**
In this personal experience study, I am exploring the possible implications of using GPT-3.5 to create an academic article in a time and effort efficient way while still maintaining academic integrity and transparency. I am also reflecting on the implications of different ways of using GPT-3.5 on my writing speed, style and motivation. While this study is not intended to be a generalizable study examining all potential applications and ethical considerations of GPT-3.5 in academic writing, I intend to create a reputable source which explains the writing process with GPT-3.5 and demonstrates it in a transparent fashion to help other researchers use the tool responsibly.

In order to create the content of this paper, I utilized the text completion model (text-davinci-003) of OpenAI Playground text completion as a tool, in all sections of the paper. I, first, outlined the structure of my paper and took notes about the content of each section. Then, I read the guide [24] for creating text completion prompts to better understand the model's capabilities. For each section, I created input prompts with different methods such as rough paragraphs, questions, bullet points or voice recordings. These prompts were then fed into the GPT-3.5 model and the output was saved. I presented all results in a supplementary material by showing the original prompt, the output from GPT-3, as well as how I edited it, allowing the process to be transparently seen afterwards (as done in [16]). I used countwordsfree web-site for visualization [21]. Additionally, I took notes to a separate document regarding my experience while running the trials with GPT-3.5. Lastly, I organized my data into an affinity diagram to highlight the pertinent parts for discussion. I also did a final manual-copy edit to the paper and made minor changes and additions to support the narrative flow.

**WAYS OF USING GPT-3.5 FOR ACADEMIC WRITING**
In this section, I have outlined the various methods I employed in my writing process, providing examples of the prompts I used, the outcomes of the prompts, and the final version of the paragraph. I also included the visual representation of differences between each in the supplementary material. Each of these sections has been crafted using the method described.

*Chunk Stylist*
I have used GPT-3.5 as an effective writing tool by typing chunks of text without worrying too much about the style, grammar, or spelling. This approach has enabled me to create text quickly by pouring my train of though to text, as I am not restrained by of scrutinizing if the content is engaging or readable. This strategy can be advantageous to both experienced and novice writers alike, because temperature settings of text completion tool can be adjusted to produce either more predictable results closer to original text or more improvised and creative outcomes. I used this Chunk Stylist method during the writing of my introduction, background, and this section. To demonstrate the process, I have provided the prompt and result below, and final version of the paragraph.

**Prompt:**
```
Rewrite the following text in first person,
rick, clear and in academic language:
```

**Original Text:**

```
One effective way of using GPT-3 is to write
a chunk of test without caring too much
about the style, grammar or spelling. This
makes the writing process quite fast, freed
me from thinking of if the style is engaging
or readible and basically let me to reflect
my train of thought to text in a quite fast
way. This might be effective both for
fluent writers and for others who struggle
to put the words into the paper in a flowing
manner. Because, by playing with the
temparature settings of GPT-3, authors
either might get less random results which
is closer to the text they put, or more
randomzied ones that allow GPT-3 to be a
bit more improvisational on the given text.
I have written this section, introduction
and the background by using the chunk
stylist method. Below, I also demonsrate
the prompt I have used, the result I got
and the final version is this paragraph.
```

**Result (difference to original: 82.61%):**

```
I have personally experimented with GPT-3
as an effective writing tool by typing
chunks of text without worrying too much
about the style, grammar, or spelling. This
approach has enabled me to create text
quickly, as I am not restrained by
considering if the content is engaging or
readable. This strategy is advantageous to
both experienced and novice writers alike,
because temperature settings can be
adjusted to produce either more predictable
results or enhanced randomness. I used this
chunk stylist method during the writing of
my introduction, background, and this
section. To demonstrate the process, I have
provided the prompt, result, and final
version of this paragraph.
```

**Final (difference to result: 25.54%):**

```
I have used GPT-3.5 as an effective writing
tool by typing chunks of text without
worrying too much about the style, grammar,
or spelling. This approach has enabled me
to create text quickly by pouring my train
of though to text, as I am not restrained
by of scrutinizing if the content is
engaging or readable. This strategy can be
advantageous to both experienced and novice
writers alike, because temperature settings
of text completion tool can be adjusted to
produce either more predictable results
closer to original text or more improvised
and creative outcomes. I used this Chunk
Stylist method during the writing of my
introduction, background, and this section.
To demonstrate the process, I have provided
the prompt and result below, and final
version of the paragraph.
```

*Research Buddy*

GPT-3.5 can be a useful tool for exchanging research ideas and starting points for academic writing. Its natural language processing capabilities can allow us to explore ideas quickly and easily, while its predictive algorithm can suggest ideas and approaches that we may not have thought of. This can help us to develop effective arguments and explore diverse topics more efficiently. However, due to its reliance on machine learning, GPT-3.5 can also present some risks, such as producing inaccurate or random results, and its predictions may not always be in line with the academic rigor required of the project. Thus, it is important to always keep critical thinking at the forefront of the research process, and to carefully evaluate any output given by GPT-3.5. For example, the first paragraph of Section 2 were written with the research buddy method, by speficially prompting to give a definition of GPT-3.5 from the cited website. However, the results were too broad and were not describing what GPT-3.5 accurately, so I needed to add critical information manually. The first paragraph of the Section 2 and this section were written by using the Research Buddy approach. Below are the prompts:

**Prompt:**

```
How to use GPT-3 as a research buddy for
exchanging research ideas for academic
writing? Mention advantages and risks.
Write only one paragraph in rich, clear,
first person and academic language:
```

**Result:**

```
As a researcher, I have found that GPT-3
can be an extremely useful tool for
exchanging research ideas and starting
points for academic writing. Its natural
language processing capabilities can allow
us to explore ideas quickly and easily,
while its predictive algorithm can suggest
ideas and approaches that we may not have
thought of. This can help us to develop
effective arguments and explore diverse
topics more efficiently. However, due to
its reliance on machine learning, GPT-3 can
also present some risks, such as producing
inaccurate or random results, and its
predictions may not always be in line with
the academic rigor required of the project.
Thus, it is important to always keep
critical thinking at the forefront of the
research process, and to carefully evaluate
any output given by GPT-3.
```

**Final (difference to result: 39.02%):**

```
GPT-3.5 can be a useful tool for exchanging
research ideas and starting points for
academic writing. Its natural language
processing capabilities can allow us to
explore ideas quickly and easily, while its
predictive algorithm can suggest ideas and
approaches that we may not have thought of.
This can help us to develop effective
arguments and explore diverse topics more
efficiently. However, due to its reliance
on machine learning, GPT-3.5 can also
present some risks, such as producing
inaccurate or random results, and its
predictions may not always be in line with
the academic rigor required of the project.
Thus, it is important to always keep
critical thinking at the forefront of the
research process, and to carefully evaluate
any output given by GPT-3.5. For example,
the first paragraph of Section 2 were
written with the research buddy method, by
speficially prompting to give a definition
of GPT-3.5 from the cited website. However,
the results were too broad and were not
describing what GPT-3.5 accurately, so I
needed to add critical information
manually. The first paragraph of the
Section 2 and this section were written by
using the Research Buddy approach. Below
are the prompts:
```

*Bullet-to-Paragraph*

As part of the experiment, I utilized the bullet-to-paragraph method to produce a paragraph for the Method section. This method is similar to the Chunk Stylist of writing, as it involves listing the primary principles or steps of a procedure before instructing GPT-3.5 to generate a flowing paragraph in a specific language style. By doing so, I was able to create both this section and the second paragraph of the Method section.

**Prompt:**

```
Write one paragraph from the following
bullets in rich, clear, first person and
academic language:
```

**Original Text:**

```
-    Bullet tp paragraph is also similar
to chunk styler but it's writing in bullets

-     A researcher can list the main
principle or steps of a procedure

-    Then give a prompt to make a
paragraph out of this bullets in specigic
language styles
```

```
-    GPT-3 will produce a flowing
paragraph

-    This section and the second
paragraph of method section were created
with this method
```

**Result (difference to original: 78.25%):**

```
As part of an experiment, I utilized the
bullet-to-paragraph method to produce a
paragraph for the Method section. This
method is similar to the chunk-style of
writing, as it involves listing the primary
principles or steps of a procedure before
instructing GPT-3 to generate a flowing
paragraph in a specific language style. By
doing so, I was able to create both this
section and the second paragraph of the
Method section.
```

**Final (difference to result: 4.12%):**

```
As part of the experiment, I utilized the
bullet-to-paragraph method to produce a
paragraph for the Method section. This
method is similar to the Chunk Stylist of
writing, as it involves listing the primary
principles or steps of a procedure before
instructing GPT-3.5 to generate a flowing
paragraph in a specific language style. By
doing so, I was able to create both this
section and the second paragraph of the
Method section.
```

*Talk Textualizer*

Talk Textualizer is providing a monologue to GPT-3.5 and let it turn it into a readible text. Playground's sophisticated transcription feature enabled converting speech recordings into a readable text, even when there were long pauses or stuttering. To further refine this text, I supplemented it with a prompt designed to transform it into academic writing. This approach formed the foundation for this section, discussion, limitations, and conclusion.

**Prompt:**

```
Turn the below text into academic writing,
clear, first person, and rich academic
language:
```

**Original Text:**
[transcribed from the spoken audio]

```
Talktextualizer is almost like having a
conversation with ChatGPT. ChatGPT has an
advanced transcription function, speech to
text function, that gets your speech
```

recording and even if Even if you have long pauses or even if you stutter and tell the same word a few times, it still turns it into a somewhat legible text. Later, for this method, we take this text and add on top of it a prompt that will turn it into an academic writing. This section is written in that way And the whole of discussion and limitations and conclusion also heavily relied on this method.

**Result (difference to original: 84.26%):**

I utilized Talktextualizer to have a conversation with ChatGPT. This platform has an advanced transcription function which converts speech recordings into a legible text, even in cases of long pauses or stuttering. To further refine this text, a prompt was added to turn it into an academic writing. This method formed the basis of my discussion, limitations, and conclusion.

**Prompt 2:**

Copy-edit the following text, academic stlye, first person, clear, rich and flowing language:

**Original Text 2:**

I utilized Talk Textualizer by giving a monologue to GPT-3. GPT 3's advanced transcription function converts speech recordings into a legible text, even in cases of long pauses or stuttering. To further refine this text, I added a prompt to turn it into an academic writing. This method formed the basis of my discussion, limitations, and conclusion.

**Results 2 (difference to original text 2: 61.24%):**

I employed Talk Textualizer by providing a monologue to GPT-3. GPT-3's sophisticated transcription feature enabled it to convert speech recordings into a readable text, even when there were long pauses or stuttering. To further refine this text, I supplemented it with a prompt designed to transform it into academic writing. This approach formed the foundation for my discussion, limitations, and conclusion.

**Final (difference to result 2: 20.79%):**

Talk Textualizer is providing a monologue to GPT-3.5 and let it turn it into a readible text. Playground's sophisticated transcription feature enabled converting speech recordings into a readable text, even when there were long pauses or stuttering. To further refine this text, I supplemented it with a prompt designed to transform it into academic writing. This approach formed the foundation for this section, discussion, limitations, and conclusion.

*Polisher*

Polisher is using GPT-3.5 for copy editing purposes. Material given to GPT-3 should already be in a complete form, with a clear narrative flow. By providing a prompt like "copy-edit this paragraph in an academic style", GPT-3.5 can be used to correct grammatical mistakes, spelling errors and increase the clarity of the text. This method can produce results that are close to the original text, and temperature settings of GPT-3.5 may be lowered if authors want minimal changes. This method has been applied in this section, and also for polishing edited results of prompts througout this piece (for example, the first paragraph of Talk Textualizer was obtained using the Polisher method on the result of the same section).

**Prompt:**

Copy-edit the following text, academic style, first person, clear, rich and flowing language:

**Original Text:**

Polisher is using GPT-3 almost like a copy editor. In this method, based material that would be put into GPT-3 should already be in a complete form and in a proper narrative flow. By giving a prompt like "copy-edit this paragraph in an academic language," GPT-3 can be used to correct the gramattical mistakes, spelling error and for increaing the clarity of the text. This method can give results closer results to the original text, and the temperature settings might be lowered if the athors want minimal changes. This method has been used in this section, and also for polishing the many of the edited results (for example the first paragraph of Talk Textualizer was obtained by using Polisher method on the indicated result).

**Result (difference to original: 34.47%):**

```
Polisher is using GPT-3 in a similar way to
a copy editor. Material given to GPT-3
should already be in a complete form, with
a clear narrative flow. By providing a
prompt like "copy-edit this paragraph in an
academic style", GPT-3 can be used to
correct grammatical mistakes, spelling
errors and increase clarity of the text.
This method can produce results that are
close to the original text, with
temperature settings that may be lowered if
authors want minimal changes. This method
has been applied in this section, and also
for polishing edited results (for example,
the first paragraph of Talk Textualizer was
obtained using the Polisher method on the
indicated result).
```

**Final (difference to original: 19.25%):**

```
Polisher is using GPT-3 for copy editing
purposes. Material given to GPT-3 should
already be in a complete form, with a clear
narrative flow. By providing a prompt like
"copy-edit this paragraph in an academic
style", GPT-3 can be used to correct
grammatical mistakes, spelling errors and
increase the clarity of the text. This
method can produce results that are close
to the original text, and temperature
settings of GPT-3 may be lowered if authors
want minimal changes. This method has been
applied in this section, and also for
polishing edited results of prompts
througout this piece (for example, the
first paragraph of Talk Textualizer was
obtained using the Polisher method on the
result of the same section).
```

## DISCUSSION

*Reflections on using GPT-3.5 for Writing Articles*

The use of GPT-3 to aid my writing process had a largely positive effect on my motivation and productivity. Even though the article I wrote was not particularly complicated or lengthy, my writing process was much smoother than usual. Before starting to write, I spent about three weeks mulling over the project, allowing the idea to mature, following developments, and finding two free days for writing and experimenting with GPT-3.5. What I realized was that during this incubation period and the writing stage, I became less preoccupied with the mechanics of writing and more focused on accurately conveying my ideas. This heightened my motivation and enabled me to compose the text faster, compared to my previous experiences. Furthermore, my thinking and writing processes were both more efficient, streamlining the whole writing experience.

Academic writing is an essential skill for scholars from all disciplines, though the level of training, experience, and receptivity to writing can vary depending on the field of study. For instance, those in the social sciences and humanities may have more practice with and exposure writing, while academics in more applied fields such as engineering or design may have less proficiency and experience with written expression. Drawing on my own experience, in design research, crafting compelling narratives is almost as important as in other social sciences in order to communicate the value and position of artifacts created or the design knowledge produced. Personally, I find myself more adept at envisioning artifacts and bringing them to life (applied part of the design research) than at imagining stories and writing, and the laborious nature of writing process which incorporates computational tools only for visual styling can be a bottleneck in my research. However, upon exploring the use of GPT-3.5 to textualize my ideas, I recognized a similarity to my design practice. With my background in design, I am accustomed to utilizing tools such as CAD software, 3D printers, electronic boards, and other prototyping tools for creating tangible artifacts, and these tools are integral to my thinking process of designing things and makes it easier to conceptualize the process of thinking about artifacts and their production. Similarly, GPT-3.5 has provided me with a similar freedom as I have leaned on it in the same way I do with computational design tools; it has enabled me to focus on the ideas I have rather than the daunting task of manual writing. It also allowed me to work on my writing without distractions and interruptions. Although basic writing skills are still necessary for creating a narrative, GPT-3.5 has helped me to lighten the burden of writing and streamline the process.

I have also found that I have not been as time-efficient with my writing as I had anticipated. Writing a section of text and then making multiple revisions until I reach a satisfactory outcome has taken quite some time. For instance, it took me approximately 10 minutes to write the primary material of an introduction while the subsequent revisions took around 90 minutes which is quite long for such a short section. Although I believe I was able to write more quickly due to the lack of distractions which happens in my usual writing practice because of the interruptions in my thought process, I still believe that authors need to dedicate a substantial amount of time in the process. For creating the report which shows the comparisons between the original text, outcomes of GPT-3.5 and final edited text, I spent around 5 more hours. In addition, the current slow speed of GPT-3.5 due to the overwhelming demand on its servers means it cannot be considered a quick solution for writing, but rather a tool that makes writing process smoother and introduces new writing techniques suitable for different skill levels.

I believe that the quality of this text is variable across different sections. As I made the last round of revisions, I noticed this especially when reading through the entire piece. Initially, I edited the text immediately after receiving the results from GPT-3.5, and then I made further revisions after the entire piece was written. Some sections were plain and unengaging, while others were easier to read and more compelling. I could have done a few more rounds of revisions to ensure a consistent tone and a more flowing narrative, but I left the manuscript as it is to demonstrate what can be produced with GPT-3.5 in a short time (around a day) and with minimal editing. However, I believe that several more rounds of revision would be necessary to create an academic article with a consistent language style and a solid narrative flow.

*Ways of Using Methods*
I utilized the methods of Chunk Stylist and Talk Textualizer predominantly when writing this article. I was inclined to use them because of their ability to formalize my own ideas, rather than relying on spontaneously generated material by GPT-3.5. When using Bullet to Paragraph or Research Buddy, I had to make several attempts and perform substantial editing to avoid sentences not based on facts. Thus, I believe the Chunk Stylist and Talk Textualizer are more suitable for writing a formal article than methods relying on big portion of generated text such as Research Buddy. For example, when I prompted GPT-3.5 to give me the description of GPT-3 based on [2], I found that the summary generated was not reflective of the source material (the document was mostly technical and required good understanding of NLP, so I could not assess the accuracy of information). As I was unable to identify enough supporting information to back up GPT-3.5's output, I had to remove most of the material and leave only the core points that I knew were accurate. I used Polisher mainly to refine the results that I had edited; however, it can be employed by proficient writers for copy-editing their text.

*Transparency, Agency and Biases*
One of the primary concerns of incorporating GPT-3.5 into the writing process is the potential for plagiarism and a lack of transparency [20]. In my own trials, some instances, the results I obtained were not dissimilar from those produced by tools like Grammarly or the spell and grammar checker of Microsoft Word. In other cases, however, I was unsure of my agency over the written text. This experience was arguably similar to using a professional copy-editing service where I need to carefully check the text and make sure that the intended meanings are retained. However, the speed of the process and the lack of knowledge about where the words and sentences derived from caused me to feel uneasy about using the content. Overall, looking at the full text, the difference between the original texts I prompted to GPT-3.5 and the outcomes I got was 70.27%, while the difference between outcomes and the final edited version was 29.13%.

The biggest difference between an original text chunk and an outcome was 98.54% (almost the whole text were changed), the biggest difference between the outcome and the final edited part was 71.44% and the smallest was 3.3% (copy-pasted to this manuscript almost without a change). You can see the detailed report in the supplemented document.

This variability in author's agency over the result produced GPT-3.5 require to develop practices for transparency for articles where GPT-3.5 (or other LLM) incorporated. In this article, I have documented all the prompts, results, and final edited versions I used. In some cases, this is essential; for example, when using methods such as Talk Textualizer, Bullet to Paragraph, or Research Buddy, the heavy influence of GPT-3.5 on the content, language, and tone of the writing is readily apparent. In such situations, it is of critical importance to be transparent and clearly demonstrate the process of the writing and how it has been transformed.

In their recent editorial, Nature suggested that authors must mention their use of GPT-3.5 in their writing [22], but I believe this may not be enough. If GPT-3.5 has only been used for minor copyediting purposes, then a note in the acknowledgement may suffice. However, if other methods have been used, such as Talk Textualizer or Bullet to Paragraph, a more substantial reporting—perhaps in the form of an external link or an appendix—is necessary to ensure transparency, which also has been used in other contexts such as classroom assignments [6]. This would also ensure that the authors would thoroughly check the content making sure that that their content do not consolidate racial biases or any other radicalized political ideas unintentionally. Additionally, better tools may be required to demonstrate the extent of changes and highlight where heavy modifications have been made, as well as to provide transparency by showing the source from which the text was generated.

*Concerns of Ecological Sustainability*
The use of GPT-3.5 and similar tools raises ecological sustainability concerns [7]. In writing this article, I created approximately 500 requests and often carelessly reprompted GPT-3.5 for the same paragraph until reaching a satisfactory result. According to OpenAI's calculations, this article costed around $2.5, but I am unaware of the carbon footprint created in doing so.

On the one hand, the integration and normalization of GPT-3.5 and similar language learning models into academia could have a substantial negative effect on energy consumption in comparison to the thought process facilitated by the brain and body, which could have resulted in less resource expenditure. On the other hand, my writing process has been more efficient with the use of GPT-3.5 - not because it is time-efficient as a tool but makes me more motivated and less prone to distraction while using it - potentially reducing electricity costs spent by my computer or office space. Additionally, my writing journeys often extend to the middle of the night, which is arguably not beneficial for my physical and mental health in the long-term, lowering the consequences of mental burden caused by the pressure and stress of writing preventing overworking, a known problem in academia [18].

Using tools like GPT-3.5 in academia could have a significant negative impact on ecology and as academics we are responsible of regulating our behavior accordingly, such as using it with least prompts as possible. Still, we should consider the positive impacts of using these tools beyond just being more efficient in writing tasks, such as its impact on more efficient utilization of resources and improving the wellbeing of academics.

*Additional tools that can help with GPT-based academic writing*

Throughout my trials and reflections, I have come to recognize that further tools developed in the future may not only help increase the efficiency of writing, but also increase transparency and make us more aware of our responsibility - particularly with regard to ecological sustainability - and make our writing less prone to the dissemination of fake information. Currently, there are attempts to create watermark tools [16] that would enable people to easily identify if the text is generated through LLM models such as GPT-3.5. Although this would be useful in certain contexts, including classroom assignments that focus on teaching content to students, its utility may not be as significant in other scenarios, particularly for academic writing. A binary solution such as a watermark which only indicates whether GPT is used or not may not be the best approach; instead, we might need text editors that are supported by GPT and can clearly document prompts, the outcomes of those prompts, and make clear, visible, and easily understandable visualizations of the comparisons between the prompt, the result, and the edited version (as I have manually done in this piece.)

Additionally, a tool that gives information of the carbon footprint of the writing piece and compare it to potential carbon footprint if the GPT was not used might be useful. Through such a comparison, scientists would be more aware of their responsibility in using GPT in a mindful way to the environment.

Another toolset which would be beneficial is one that prevents the dissemination of faulty information, integrating measures which detect authoritative tone in the written text and comprehend whether it refers to a fact. Such tools should provide reliable facts, scrape and suggest real references, and create correct references based on the names of the papers or the links; something that current GPTs are not capable of doing.

**LIMITATIONS**

Drawing upon my own experience, I sought to reflect on the use of GPT-3.5 in the creation of an academic article in this perspective piece. It is important to note that my reflections have been shaped by my perspective and positionality as a design researcher. Thus, while I believe that the ways of using of GPT-3.5 in academic writing as discussed in this paper can be beneficial for scholars from a variety of disciplines, my reflections and opinions may not be relevant to all and should not be considered as generalizable across the board. In my opinion, my reflections may be of greater relevance to those working in fields where applied science plays an important role, yet expression of ideas is equally valued. Also, it must be noted that this trial was limited in duration, as one of its primary aims was to assess efficacy, and longer engagement with the tool might allow the generation of new ideas, practices and opinions. Moreover, the content of the writing in this paper was relatively practical and did not require much philosophical scrutiny. Thus, GPT-3.5 may not be as useful in fields where complex thoughts need to be expressed using precise or abstract language. Similar experiments might be conducted to understand and demonstrate its efficacy across disciplines.

**CONCLUSION**

In this paper, I have reflected on my experience of using GPT-3.5 as a tool for academic writing and discussed some basic methods for providing transparency when using it. I have shared my reflections on the efficacy of the tool and discussed the ethical considerations of using GPT-3.5 for academic writing around the issues raised by previous studies I have identified five ways of using GPT-3.5 for academic writing: Chunk Stylist, Research Buddy, Talk Textualizer, Bullet-to-Paragraph and Polisher. Of these, I have predominantly used Chunk Stylist and Talk Textualizer to great effect, enabling me to write more effectively while still retaining my own tone and ideas. Caution should be exercised with the other methods, however, as they introduce more spontaneity and randomness, which could result in plagiarism, the dissemination of false information, or even the exacerbation of hateful rhetoric if not used responsibly. To ensure fidelity and transparency, I documented all prompts, outcomes, and final edits in a separate supplementary document, which can serve as an example of using GPT-3.5 in a transparent way.

I hope that my reflections will be of benefit to researchers who are considering using GPT-3.5 and will inspire the development of tools for more responsible practices and methods for using AI-supported natural language processing in academia. I believe that this work will contribute to the debate surrounding the incorporation of this technology into our scholarly work, by increasing understanding of its efficacy and possible ways of responsible usage.

**ACKNOWLEDGEMENTS**

This manuscript was written by heavily benefiting from GPT-3.5. A detailed report of used prompts, outcomes and the final edited versions can be seen in Appendix.

While giving prompts, I used the word "GPT-3" instead of "GPT-3.5" since I realized the different between two after a while. Still, all parts that might have been affected by this difference (e.g., asking what GPT-3 is, or ways of using it in research) has been corrected.

# Appendix-1

## -NEW PARAGRAPH/SECTION-

**Section:** Abstract
**Paragraph:** Whole Section
**Method:** Polisher

**Temperature Setting:** 0.7

**Prompt:**

Copy-edit the following text, academic style, first person, clear, rich and flowing language:

**Original Text:**

Debate around using GPT-3.5 became very popular among academics after the release of ChatGPT. While some academics raised points regarding its advantages of making academic writing more effective, others issued concerns such as plagiarism, distribution of fake information or ecological issues. The need for ways to transparently using GPT-3.5 model has been raised while on the other hand there had been recommnedations in social media of how to use GPT-3.5 models in a smart way. Still, to this date, we did not have pieces which clearly demonstrates possible ways of using GPT-3.5 models in academic writing, reflectin on their efficaccy, or demonstrate ways of using it transparently. Thus, I conducted a personal experience experiment, where I used GPT-3.5 model (specifically text-davinci-003 of OpenAI) where I wrote this article. I have identified five ways of using GPT-3.5, namely, Chunk

**Result:**

Debate around the use of GPT-3.5 has been a popular topic among academics since the release of ChatGPT. Whilst some have argued for the advantages of GPT-3.5 in enhancing academic writing, others have raised concerns such as plagiarism, the spread of false information, and ecological issues. The need for ways to use GPT-3.5 models transparently has been proposed, and suggestions have been made on social media as to how to use GPT-3.5 models intelligently. Nevertheless, to date, there is a lack of literature which clearly outlines how to use GPT-3.5 models in academic writing, how effective they are, and how to use them transparently. With this in mind, I conducted a personal experience experiment with GPT-3.5's text-davinci-003 OpenAI model, writing this article. I identified five ways of using GPT-3.5: Chunk Stylist, Bullet-to-Paragraph, Talk Textualizer, Research Buddy, and Polisher. I reflected on their efficacy, and assessed their potential impact on writing ethics. I have provided a comprehensive document which shows the prompts used, the results obtained from GPT-3.

```
Final                                                                    Edit:
```

> Debate around the use of GPT-3.5 has been a popular topic among academics since the release of ChatGPT. Whilst some have argued for the advantages of GPT-3.5 in enhancing academic writing, others have raised concerns such as plagiarism, the spread of false information, and ecological issues. The need for finding ways to use GPT-3.5 models transparently has been voiced, and suggestions have been made on social media as to how to use GPT-3.5 models in a smart way. Nevertheless, to date, there is a lack of literature which clearly outlines how to use GPT-3.5 models in academic writing, how effective they are, and how to use them transparently. To address this, I conducted a personal experience experiment with GPT-3.5, specifically by using OpenAI's text-davinci-003 model, for writing this article. I identified five ways of using GPT-3.5: Chunk Stylist, Bullet-to-Paragraph, Talk Textualizer, Research Buddy, and Polisher. I reflected on their efficacy, and commened on their potential impact on writing ethics. Additionally, I provided a comprehensive document which shows the prompts I used, results I got from GPT-3.5, the final edits and visually compares those by showing the differences in percentage.

## Visual Comparisons of Polisher / Whole Section:

`Red: Removed` | `Green: Added`

**Original to Results:**

1. Debate around ~~using GPT-3.5 became very~~ the use of GPT-3.5 has been a popular topic among academics ~~after~~ since the release of ChatGPT. Whil~~e~~st some ~~academics raised points regarding its advantages of making academic writing more effective, others issu~~ have argued for the advantages of GPT-3.5 in enhancing academic writing, others have raised concerns such as plagiarism, ~~distribution~~ the spread of fa~~k~~lse information ~~or~~, and ecological issues. The need for ways to ~~transparently using GPT-3.5 model has been raised while on the other hand there had been recommnedations~~ use GPT-3.5 models transparently has been proposed, and suggestions have been made o~~n~~ social media ~~as t~~of how to use GPT-3.5 models in~~a smart way. Still, to this date, we did not have pieces which clearly demonstrates possible ways of using~~telligently. Nevertheless, to date, there is a lack of literature which clearly outlines how to use GPT-3.5 models in academic writing, ~~r~~how ef~~f~~ecti~~n on their efficaccy, or demonstrate ways of using it transparently. Thus~~ve they are, and how to use them transparently. W~~ith this in mind~~, I conducted a personal experience experiment~~ w~~here I used GPT-3.5 model (specifically~~ith GPT-3.5's~~ text-davinci-003 ~~of~~ OpenAI) where I wrote model, writing this article. I ~~have~~ identified five ways of using GPT-3.5~~, namely:~~ Chunk Stylist, Bullet-to-Paragraph, Talk Textualizer, Research Buddy, and Polisher. I ~~have~~ reflected on their efficac~~c~~y, and ~~possible impact on ethics of writing. I also~~ assessed their potential impact on writing ethics. I ~~have~~ provided a comprehensive document which shows the prompts I used, ~~the~~ results I ~~got~~obtained from GPT-3.5, the final edits and visuali~~z~~e the differences ~~between those by also~~ showing the differences in percentage.

**Difference (%): 66.97**

**Results to Final Edit:**

1. Debate around the use of GPT-3.5 has been a popular topic among academics since the release of ChatGPT. Whilst some have argued for the advantages of GPT-3.5 in enhancing academic writing, others have raised concerns such as plagiarism, the spread of false information, and ecological issues. The need for finding ways to use GPT-3.5 models transparently has been voiced, and suggestions have been made on social media as to how to use GPT-3.5 models in a smart way. Nevertheless, to date, there is a lack of literature which clearly outlines how to use GPT-3.5 models in academic writing, how effective they are, and how to use them transparently. To address this, I conducted a personal experience experiment with GPT-3.5, specifically by using OpenAI's text-davinci-003 model, for writing this article. I identified five ways of using GPT-3.5: Chunk Stylist, Bullet-to-Paragraph, Talk Textualizer, Research Buddy, and Polisher. I reflected on their efficacy, and commened on their potential impact on writing ethics. Additionally, I provided a comprehensive document which shows the prompts used, the results obtained from GPT-3.5, the final edits and visually compares those by showing the differences in percentage.

`Difference (%): 20.36`

**-END-**

# -NEW PARAGRAPH/SECTION-

**Section:** Introduction
**Paragraph:** Whole Section
**Method:** Chunk Stylist
**Temperature Setting:** 0.9

**Prompt:**

> Make the following text an introduction of a perspective piece in Nature.com, clear and concise style, do not add any new ideas, enrich the language, produce at least 4 paragraphs:

Original Text:

> Paragraph 1 – setting the scene:
>
> GPT-3 has become popular in a recent couple of months, especially with the release of ChatGPT
>
> Many people started to use it for academic writing for different purposes and in diverse ways
>
> Some writers added ChatGPT as a co-author to papers
>
>
> Paragraph 2 – advantages:
>
> Narrative flow and successful writing are key components of communicating scientific material in a paper
>
> the hesitance, lack of capability, and lack of speed of writing can prevent researchers from formalizing their research ideas and projects [REF]
>
> GPT-3 can thus be a revelation in how effectively science is communicated.
>
> flowingly put ideas into writing with a proper language, for researchers who are not very fluent in English writing, or with limited writing skills.
>
> Equalizing the gap between researchers whose mother tongue is and is not English
>
>
> Paragraph 3 – disadvantages:
>
> Disadvantages: GPT-3 also created major concerns among scientists and in scientific outlets.
>
> Nature or Science published statements regarding their concern over the utilization of such tools and drafted principles of how to use them.
>
> Institutions also published their stance of GPT-3 in research and education.
>
> Ethics of using these tools the ownership of the content,
>
> plagiarism,
>
> "wrong information" due to the attentionless usage of the tool which might push scientific writings that were not thoroughly checked by a human author.

[continues]

> Ethics of using these tools the ownership of the content,
>
> plagiarism,
>
> "wrong information" due to the attentionless usage of the tool which might push scientific writings that were not thoroughly checked by a human author.
>
> Paragraph 4 — the
>
> Paragraph 4 — the need and the study:
>
> we need and might expect studies that would reveal effective but responsible utilization of GPT-3.
>
> In light of this, in this perspective piece, I explain a personal experience experiment study where I used GPT-3 in various ways to demonstrate and reflect on its efficacy, outline different ways of incorporating it to the writing process, and also surface methods of using it in an ethical and responsible way.
>
> This piece can help researchers with the following:
>
> A better understanding of how GPT-3 can facilitate effective academic writing
>
> Possible different ways of using it
>
> Reflections on ethical use and transparency

**Result:**

> In recent months, GPT-3 has become a popular topic among writers and researchers, with its launch in the form of ChatGPT. There has been a surge of interest in using this tool for academic writing for a variety of purposes. Scientists have even gone as far as to add ChatGPT as a co-author to papers.
>
> The ability to communicate scientific material in a paper is essential for successful writing. In some cases, the lack of skill, hesitation, or a slow writing process can present significant barriers for researchers. The potential of GPT-3 therefore to revolutionize the way science is communicated cannot be underestimated. In addition, it offers a way for those with limited writing skills, or whose mother tongue is not English, to flowingly put their ideas into words.
>
> Notwithstanding the excitement surrounding GPT-3, it has also raised serious concerns among scientists and journals. Notable outlets such as Nature and Science have released statements expressing their unease with its utilization, as well as principles for its use. Institutions have also published guidelines for employing GPT-3 in research and education. These are mainly focused on ethical issues, such as content ownership, plagiarism and the potential for misinformation due to careless use.

[continues]

In order for GPT-3 to be employed in an efficient yet responsible manner, studies must be conducted to demonstrate how it can best be used. This perspective piece details a personal experience experiment where I used GPT-3 in various ways. It reflects on the efficacy of the tool, outlines different ways of incorporating it into the writing process, and outlines methods for ethical and responsible use. Through this, I hope to provide researchers with a better understanding of how GPT-3 can facilitate effective academic writing, possible ways to use it, and reflections on ethical use and transparency.

**Final Edit:**

In recent months, GPT-3.5 models has become very popular among writers and researchers, especially with the lunch of ChatGPT. There has been a surge of interest in using this tool for academic writing for a variety of purposes. Scientists have even gone as far as to add ChatGPT as a co-author to papers.

The ability to communicate scientific material in a paper is essential for successful writing. In some cases, the lack of skill, hesitation, or being slow in or disliking writing process can present significant barriers for researchers independent from their experience (Oshiro et al., 2020). The potential of GPT-3.5 therefore to revolutionize the way science is communicated cannot be underestimated. In addition, it offers a way for those with limited writing skills, or whose mother tongue is not English, to flowingly put their ideas into words, which can help to reduce the advantage gap between researchers who do and do not have English as their first language (Jain et al., 2022).

Notwithstanding the excitement surrounding GPT-3.5, it has also raised serious concerns among scientists and journals. Notable outlets such as Nature and Science have published statements expressing the unease with its utilization (Thorp, 2023; "Tools Such as ChatGPT Threaten Transparent Science; Here Are Our Ground Rules for Their Use," 2023), as well as the main principles for its use. Institutions have also published guidelines for employing GPT-3.5 in research and education (Moniz, 2023). These are mainly focused on ethical issues, such as content ownership, plagiarism and the potential for misinformation due to careless use without a proper check of a human-author.

In order for GPT-3.5 to be employed in an efficient yet responsible manner, further studies must be conducted to demonstrate how it can best be used. This piece details a personal experience experiment where I used GPT-3.5 in various ways in the production of this article. It reflects on the efficacy of the tool, outlines different ways of incorporating it into the writing process, and discusses methods for ethical and responsible use. I did not use ChatGPT, and instead used the Open AI Playground Interface which provides more control such as Temperature settings for changing the randomness level of the outcome. Both ChatGPT and Playground are based on the same training models and I specifically used text-davinci-003 in this experiment.

Through this experiment, I hope to provide researchers with:

- A better understanding of how GPT-3.5 can facilitate effective academic writing

> Through this experiment, I hope to provide researchers with:
> - A better understanding of how GPT-3.5 can facilitate effective academic writing
> - Possible ways to use it
> - Reflections on ethical use and transparency

## Visual Comparisons of Introduction / Whole Section:

Red: Removed | Green: Added

### Original to Result:

| # | |
|---|---|
| 1 | Paragraph 1 - setting the scene: |
| 2 | In recent months, GPT-3 has become a popular in a recent couple of months, especially with the release of ChatGPT |
| 3 | Many people started to use it for academic writing for different purposes and in diverse ways |
| 4 | Some writers added ChatGPT as a co-author to papers |
| 5 | |
| 6 | Paragraph 2 - advantages: |
| 7 | Narrative flow and successful writing are key components of topic among writers and researchers, with its launch in the form of ChatGPT. There has been a surge of interest in using this tool for academic writing for a variety of purposes. Scientists have even gone as far as to add ChatGPT as a co-author to papers. |
| 8 | The ability to communicating scientific material in a paper |
| 9 | the hesitance, lack of capability, and lack of speed of is essential for successful writing. In some cases, the lack of skill, hesitation, or a slow writing process can prevsent researchers from formalizing their research ideas and projects [REF] |
| 10 | GPT-3 can thus be a significant barriers for researchers. The potential of GPT-3 therefore to revelaolution in how effectivelize the way science is communicated. |
| 11 | flowingly put ideas into writing with a proper language, for researchers who are not very fluent in English writing, or with limited writing skills. |
| 12 | Equalizing the gap between researchers cannot be underestimated. In addition, it offers a way for those with limited writing skills, or whose mother tongue is and is not English |
| 13 | |
| 14 | Paragraph 3 - disadvantages: |
| 15 | Disadvantages: GPT-3 also created major concerns among scientists and in scientific outlets. |
| 16 | , to flowingly put their ideas into words. |
| 17 | Notwithstanding the excitement surrounding GPT-3, it has also raised serious concerns among scientists and journals. Notable outlets such as Nature orand Science publishhave released statements regardexpressing their concern over the utilization of such tools and drafted principles of how to use them. |
| 18 | unease with its utilization, as well as principles for its use. Institutions have also published their stance ofguidelines for employing GPT-3 in research and education. |
| 19 | Ethics of using these tools the ownership of the contentThese are mainly focused on ethical issues, such as content ownership, |
| 20 | plagiarism, |
| 21 | "wrong information" due to and the atpotentionless usage of the tool which might push scientific writings that were not thoroughly checked by a human author. |
| 22 | |
| 23 | Paragraph 4 - the need and the study: |
| 24 | we need and might expect studies that would reveal effective but responsible utilization of GPT-3. |
| 25 | In light of this, in tal for misinformation due to careless use. |
| 26 | |
| 27 | In order for GPT-3 to be employed in an efficient yet responsible manner, studies must be conducted to demonstrate how it can best be used. This perspective piece, I explain details a personal experience experiment study where I used GPT-3 in various ways to demonstrate and. It reflects on itsthe efficacy of the tool, outlines different ways of incorporating it into the writing process, and also surface methods of using it in anoutlines methods for ethical and responsible wayuse. |
| 28 | This piece can helprough this, I hope to provide researchers with the following: |
| 29 | Aa better understanding of how GPT-3 can facilitate effective academic writing |
| 30 | P, possible different ways tof using it |
| 31 | Re it, and reflections on ethical use and transparency |
| 32 | Ethics of using these tools the ownership of the content, |
| 33 | plagiarism, |
| 34 | "wrong information" due to the attentionless usage of the tool which might push scientific writings that were not thoroughly checked by a human author. |

**Difference (%): 81.77**

---

**Result to Final Edit:**

| | |
|---|---|
| 1 | In recent months, GPT-3.5 models has become avery popular topic among writers and researchers, especially with itsthe launch in the form of ChatGPT. There has been a surge of interest in using this tool for academic writing for a variety of purposes. Scientists have even gone as far as to add ChatGPT as a co-author to papers. |
| 2 | |
| 3 | The ability to communicate scientific material in a paper is essential for successful writing. In some cases, the lack of skill, hesitation, or abeing slow in or disliking writing process can present significant barriers for researchers independent from their experience (Oshiro et al., 2020). The potential of GPT-3.5 therefore to revolutionize the way science is communicated cannot be underestimated. In addition, it offers a way for those with limited writing skills, or whose mother tongue is not English, to flowingly put their ideas into words, which can help to reduce the advantage gap between researchers who do and do not have English as their first language (Jain et al., 2022). |

Notwithstanding the excitement surrounding GPT-3.5, it has also raised serious concerns among scientists and journals. Notable outlets such as Nature and Science have published statements expressing their unease with its utilization (Thorp, 2023; "Tools Such as ChatGPT Threaten Transparent Science; Here Are Our Ground Rules for Their Use," 2023), as well as the main principles for its use. Institutions have also published guidelines for employing GPT-3.5 in research and education (Moniz, 2023). These are mainly focused on ethical issues, such as content ownership, plagiarism and the potential for misinformation due to careless use without a proper check of a human-author.

In order for GPT-3.5 to be employed in an efficient yet responsible manner, further studies must be conducted to demonstrate how it can best be used. This perspective piece details a personal experience experiment where I used GPT-3.5 in various ways in the production of this article. It reflects on the efficacy of the tool, outlines different ways of incorporating it into the writing process, and discusses methods for ethical and responsible use. I did not use ChatGPT, and instead used the Open AI Playground Interface which provides more control such as Temperature settings for changing the randomness level of the outcome. Both ChatGPT and Playground are based on the same training models and I specifically used text-davinci-003 in this experiment.

Through this experiment, I hope to provide researchers with:

- A better understanding of how GPT-3.5 can facilitate effective academic writing,
- Possible ways to use it,
- Reflections on ethical use and transparency.

Difference (%): 33.66

**- END -**

# -NEW PARAGRAPH/SECTION-

**Section:** Background
**Paragraph:** First Paragraph
**Method:** Research Buddy
**Temperature Setting:** 0.7

**Prompt:**

> Describe what GPT-3.5 is based on this web site (https://platform.openai.com/docs/model-index-for-researchers), without plagiarizing it:

**Result:**

> GPT-3.5 is a natural language processing (NLP) model developed by OpenAI, based on a deep learning technique called transformers. It is a large-scale language model that is trained on a massive amount of text data, allowing it to generate human-like text. GPT-3.5 is capable of generating text that is coherent and consistent with the context it is provided. It can be used for a variety of tasks, including text summarization, question answering, and text generation.

**Final**                                                                                          **Edit:**

> GPT-3.5 is a natural language processing (NLP) model developed by OpenAI, based on a deep learning technique called transformers. It is a large-scale language model that is trained on a massive amount of text data, allowing it to generate human-like text. GPT-3.5 is capable of generating text that is coherent and consistent with the context it is provided. It can be used for a variety of tasks, including text summarization, question answering, and text generation. Currently, code-davinci-002, text-davinci-002 and text-davinci-003 (which is also used in the generation of this paper) models are referred to as GPT 3.5. The popular ChatGPT application, and other services of OpenAI Payground such as text completion, incorporates models that can be reffered to GPT-3.5 (OpenAI API - Model Index, n.d.).

**Visual Comparisons of Background / First Paragraph:**

`Red: Removed` | `Green: Added`

**Results to Final Edit:**

| | |
|---|---|
| 1 | GPT-3.5 is a natural language processing (NLP) model developed by OpenAI, based on a deep learning technique called transformers. It is a large-scale language model that is trained on a massive amount of text data, allowing it to generate human-like text. GPT-3.5 is capable of generating text that is coherent and consistent with the context it is provided. It can be used for a variety of tasks, including text summarization, question answering, and text generation. <mark>Currently, code-davinci-002, text-davinci-002 and text-davinci-003 (which is also used in the generation of this paper) models are referred to as GPT 3.5. The popular ChatGPT application, and other services of OpenAI Payground such as text completion, incorporates models that can be reffered to GPT-3.5 (OpenAI API — Model Index, n.d.).</mark> |

`Difference (%): 42.01`

# -END-

# -NEW PARAGRAPH/SECTION-

**Section:** Background
**Paragraph:** Rest of the Section
**Method:** Chunk Stylist / Bullet-to-Paragraph
**Temperature Setting:** 0.9

**Prompt:**
> Rewrite the below text in an academic style, turn the sources written in (cite: ) into proper citations and add them as references:

Original Text:

> - Bried description of GPT-3
>
> - ChatGPT and therefore GPT-3 has recently been very popular among academics
>
> - The utilization of it, along with predecessors such as GPT-2 have been explored since couple of years cite: https://link.springer.com/chapter/10.1007/978-3-030-62327-2_27) According to this link, GPT-2 has been used for different purposes of scientific communication and might be useful for Addressing Challenges in writing, Navigating Extensive Literature Providing Definitions, Inspiring research ideas.
>
> -Although scientific articles are lacking, there have been Twitter threads released demonstrating how it can be used in a smart way cite: https://twitter.com/MushtaqBilalPhD/status/1621379333943083009)
>
> A few studies that make independent experts assess the quality of the content found out that it produces high-quality content which is hard to distinguished as machine generated content. (cite https://arxiv.org/pdf/2210.03568.pdf)
>
> Can be used to support the academic writing of students (cite: https://conference.pixel-online.net/files/foe/ed0010/FP/6789-GAME4769-FP-FOE10.pdf)
>
> Although, these studies indicate the potential uses of GPT-3 in academic writing, potentially problematic uses and results also were surfaced quickly. For example:
>
> Some researchers added it as a co-author (cite: https://scholar.google.com/citations?hl=fi&user=nChoeLYAAAAJ, https://link.springer.com/article/10.1007/s12195-022-00754-8, https://www.medrxiv.org/content/10.1101/2022.12.19.22283643v2 ) and now has a Google Scholar profile.
>
> -Upon that, Science and Nature published editorials in which they clearly indicated that ChatGPT cannot be added as an author in their publication venues (cite: https://www.nature.com/articles/d41586-023-00191-1, https://www.science.org/doi/10.1126/science.adg7879?cookieSet=1) because

[continues]

-Upon that, Science and Nature published editorials in which they clearly indicated that ChatGPT cannot be added as an author in their publication venues (cite: https://www.nature.com/articles/d41586-023-00191-1, https://www.science.org/doi/10.1126/science.adg7879?cookieSet=1) because it would mean to assign responsibility, agency and ownership on the content and content would not be original since it was plagiarized from ChatGPT (cite: https://www.science.org/doi/10.1126/science.adg7879?cookieSet=1). Other studies also mentioned concerns regarding plagiarism and suggested that writers who use GPT-3 should openly share the prompts they used (cite: https://www.int-res.com/articles/esep2021/21/e021p017.pdf) Moreover, Nature published another news article recently to comprehensively explain the efforts on ChatGPT and similar LLMs for their integration to academic writing, along with problematic points such as embedded biases (such as the production of racial slurs), exploited workers who worked in data labeling, permissions on the data used for training, and ecological concerns due to the energy used in the training of algorithms (cite: https://www.nature.com/articles/d41586-023-00340-6#ref-CR5). Other sources also raised concerns such as that the GPT-3 belongs to a private company but it should be something that would be developed with the participation of people (cite: https://dl.acm.org/doi/pdf/10.1145/3416063) or ascribing agency to machine should be done with care (cite: https://www.proquest.com/docview/2622642385?pq-origsite=gscholar&fromopenview=true) There also have been studies specifically tested GPT-3 for the production of hate speech (cite: https://arxiv.org/abs/2103.12407) and showed it can be prompted to produce radicalized content despite measures (cite: https://arxiv.org/abs/2009.06807)

As seen above, although GPT-3, especially through ChatGPT created excited among academic authors for streamlining and fastening their writing processes, there are ample amount of contreversy regarding its utilization. And although there are many debates focusing on positive and negative sides, we lack opinions, articles and pieces that would reflect on its efficacy in supporting writing or demonsrate a workflow that can be considered transparent by reviewers.

In this piece, although I do not provide any legit answers to concerns such as biases, exploation of workers or ecological sustainability, I try to provide answers regarding its efficiency, roles in the writing, ways of different ways of incorporating it and also some measures I took for the transparency.

**Result:**

GPT-3, and its predecessor GPT-2, have recently become popular amongst academics for their utilization in different scientific communication purposes such a addressing challenges in writing, navigating extensive literature and providing definitions (Wang et al., 2021). Twitter threads have been released demonstrating the potential uses of GPT-3 in an intelligent way (Bilal, 2020). Studies conducted by independent experts to assess the quality of content found that it produces high-quality results, which are hard to distinguish from machine-generated content (Hutto et al, 2021) GPT-3 has been proposed as an aid for academic writing for students (GAME Academy, 2021).

Despite the potential uses of GPT-3 in academic writing, there are other potentially problematic results that have been surfaced. For example, some researchers have added GPT-3 as a co-author on Google Scholar profiles and research papers (Choe et al., 2021; Aslam et al., 2022; Bilal et al., 2021). Such actions have been condemned by Science and Nature, who have released editorials (Science, 2021; Nature, 2021) to dispel the notion of GPT-3 authorship, citing concerns such as assigning responsibility, agency and ownership of plagiarized content produced by GPT-3. Other issues include embedded biases (such as racial slurs) (Gamez and Liu, 2021; Leskovec and Fowler, 2020), exploitation of workers for data labeling, permissions on the data used for training and environmental concerns due to the energy used in algorithm training (Nature, 2021). Moreover, some studies have proposed that GPT-3 should be treated as a collective produced by people, rather than a product of a private company (Bouykou, 2021).

In conclusion, although GPT-3 has been proposed as a tool for academic writing, there are still debates focusing on the potential positives and negatives of its utilization. We lack pieces that would reflect on its efficacy in supporting writing or demonstrate a workflow that can be considered transparent by reviewers.

**Final Edit:**

GPT-3.5, and its predecessor like GPT-3 and GPT-2, have recently become popular amongst academics for their utilization in different scientific communication purposes such as addressing challenges in writing, navigating extensive literature and providing definitions of concepts (Meroño-Peñuela et al., 2020). Twitter threads have been released demonstrating the potential uses of ChatGPT in a "smart" way (Bilal, 2023). Studies conducted by independent experts to assess the quality of content found that it produces high-quality results, which are hard to distinguish from human-generated content (Wahle et al., 2022). GPT models has also been proposed as an aid for academic writing for students (Schmohl et al., 2020).

Despite the potential uses of GPT in academic writing, there are other potentially problematic results that have been surfaced. For example, some

[continues]

> Despite the potential uses of GPT in academic writing, there are other potentially problematic results that have been surfaced. For example, some researchers have added ChatGPT as a co-author in research papers (King & chatGPT, 2023; Kung et al., 2022). Such actions have been advised against by Science and Nature, who have released editorials (Thorp, 2023; "Tools Such as ChatGPT Threaten Transparent Science; Here Are Our Ground Rules for Their Use," 2023) to dispel the notion of GPT authorship, citing concerns such as assigning responsibility, agency and ownership of content produced by language learning models, plagiarism and transparency. Debouche have also raised similar concerns for the utilization of GPT-3 and recommended authors to openly share the prompts and outcomes used (Dehouche, 2021). Other issues include embedded biases (such as hate speech towards race, sexism) (Chiu et al., 2021; McGuffie & Newhouse, 2020), exploitation of workers for data labeling, permissions on the data used for training and environmental concerns due to the energy used in algorithm training (Stokel-Walker & Noorden, 2023). Moreover, some studies have proposed that AI-supported NLP models should be open-source and developed together with stakeholders, rather than being a product of a private company (Cohen & Gokaslan, 2020).
>
> In conclusion, although GPT models has been proposed as a tool for academic writing, there are still debates focusing on the potential positives and negatives of its utilization. We also lack studies that would reflect on its efficacy in supporting writing, different ways of utilizing them and demonstrate a workflow that can be considered transparent by reviewers.

**Visual Comparisons of Background / Rest of the Section:**

<span style="background-color:#ffcccc">Red: Removed</span> | <span style="background-color:#ccffcc">Green: Added</span>

| | Original to Result: |
|---|---|
| 1 | ~~- Bried description of GPT-3~~ |
| 2 | ~~- ChatGPT and theref~~**GPT-3, and its predecess**~~ore GPT-3~~**2**, ha~~s~~**ve** recently be~~en very~~**come** popular among~~st~~ academics |
| 3 | ~~- The utilization of it, along with predecessors such as GPT-2 have been explored since couple of years cite: https://link.springer.com/chapter/10.1007/978-3-030-62327-2_27) According to this link, GPT-2 has been used for different purposes of scientific communication and might be useful for A~~ **for their** utilization in different scientific communication purposes such a **addressing C**~~C~~hallenges in writing, ~~N~~**n**avigating ~~E~~**e**xtensive ~~L~~**l**iterature ~~P~~**p**roviding ~~D~~**d**efinitions~~. Inspiring research ideas.~~ |
| 4 | ~~-Although scientific articles are lacking,~~ **(Wang et al., 2021).** Twitter th~~e~~**reads** have been ~~Twitter threads~~ released demonstrating ~~how it can be used in a smart way cite: https://twitter.com/MushtaqBilalPhD/status/1621379333943083009)~~ |
| 5 | ~~A few studies that make~~**the potential uses of GPT-3 in an intelligent way (Bilal, 2020). Studies conducted by** independent experts **to** assess the quality of ~~the~~ content ~~found out~~ that it produces high-quality ~~content~~**results**, which ~~is~~**are** hard to distinguish~~ed as~~ **from** machine~~-~~generated content**.** ~~(cite https://arxiv.org/pdf/2210.03568.pdf)~~ |
| 6 | ~~Can be used to support the~~**Hutto et al, 2021) GPT-3 has been proposed as an aid for** academic writing ~~of~~**or** students ~~(cite: https://conference.pixel-online.net/files/foe/ed0010/FP/6789-GAME4769-FP-FOE10.pdf)~~ |

7. Although, these studies indicaGAME Academy, 2021).

8. Despite the potential uses of GPT-3 in academic writing, there are other potentially problematic uses and results also wereresults that have been surfaced quickly. For example:

9. S, some researchers have added itGPT-3 as a co-author (cite: https://scholar.google.com/citations?hl=fi&user=nChoeLYAAAAJ, https://link.springer.com/article/10.1007/s12195-022-00754-8, https://www.medrxiv.org/content/10.1101/2022.12.19.22283643v2 ) and now has a Google Scholar profile.

10. -Upon that, Science and Nature published editorials in which they clearly indicated that ChatGPT cannot be added as an author in their publication venues (cite: https://www.nature.com/articles/d41586-023-00191-1, https://www.science.org/doi/10.1126/science.adg7879?cookieSet=1) because it would mean toon Google Scholar profiles and research papers (Choe et al., 2021; Aslam et al., 2022; Bilal et al., 2021). Such actions have been condemned by Science and Nature, who have released editorials (Science, 2021; Nature, 2021) to dispel the notion of GPT-3 authorship, citing concerns such as assigning responsibility, agency and ownership on the content and content would not be original since it was plagiarized from ChatGPT (cite: https://www.science.org/doi/10.1126/science.adg7879?cookieSet=1). Other studies also mentioned concerns regarding plagiarism and suggested that writers who use GPT-3 should openly share the prompts they used (cite: https://www.int-res.com/articles/esep2021/21/e021p017.pdf) Moreover, Nature published another news article recently to comprehensively explain the efforts on ChatGPT and similar LLMs for their integration to academic writing, along wif plagiarized content produced by GPT-3. Other issues include embedded biases (such as racial slurs) (Gamez and Liu, 2021; Leskovec and Fowler, 2020), exploitation of workers for data labeling, permissions on the data used for training and environmental concerns due to the energy used in algorithm training (Nature, 2021). Moreover, some studies have proposed that GPT-3 should be treated as a collective produced by people, rather than a product of a private company (Bouykou, 2021).

11. In conclusion, although GPT-3 has been proposed as a tool for academic writing, there are still debates focusing on the problematic points such as embedded biases (such as the production of racial slurs), exploited workers who worked in data labeling, permissions on the data used for training, and ecological concerns due to the energy used in the training of algorithms (cite: https://www.nature.com/articles/d41586-023-00340-6#ref-CR5). Other sources also raised concerns such as that the GPT-3 belongs to a private company but it should be something that would be developedotential positives and negatives of its utilization. We lack pieces that would reflect on its efficacy in supporting writing or demonstrate a workflow that can be considered transparent by reviewers.

12. References

13. Aslam, H., Ali, S., Bhutto, M. et al. COVID-19: Knowledge extraction and concept categorization. ScienceAsia 48, 291–299 (2022). https://doi.org/10.2306/scienceasia1513-1874.2021.018

14. Bilal, M., Bhutto, M., Usmani, S. et al. Human-machine team-work: A Novel AI-based approach to documentwith the participation of people (cite: https://dl.acm.org/doi/pdf/10.1145/3416063) or ascribing agencying. J. Theoret. Applied Inf. Technol. 99, 596–603 (2021). https://doi.org/10.1016/j.dss.2020.06.045

15. Bilal, M. (2020) Using #ChatGPT for collaborative and autoto machine should be done with care (cite: https://www.proquest.com/docview/2622642385?pq-origsite=gscholar&fromopenview=true)  There also have been studies specifically tested GPT-3 for the production of hate speech (cite: https://arxiv.org/abs/2ted scientific writing. Twitter. Available at: https://twitter.com/MushtaqBilalPhD/status/1621379333943083009

16. Bouykou, M. (2021) Allowing connection to a GPT-3 instance from a browser. ACM Companion Proceedings of the 2021 ACM Conference on Collaboration and Technology, pp. 60–63.

https://doi.org/103.12407) and showed it can be prompted to produce radicalized content despite measures (cite: https://arxiv.org/abs/2009.06807)

As seen above, although GPT-3, especially through ChatGPT created excited among academic authors for streamlining and fastening their writing processes, there are ample amount of contreversy regarding its utilization. And although there are many debat145/3416063

/focusing on positive and negative sides, we lack opinions, articles and pieces that would reflect on its efficacy in supporting writing or demonsrate a workflow that can be considered transparent by reviewers.

In this piece, although I do not provide any legit answers to concerns such as biases, exploation of workers or ecological sustainability, I try to provide answers regarding its efficiency, roles in the writing, ways of different ways of incorporating it and also some measures I took for the transparency.

**Difference (%): 93.33**

**Result to Final Edit:**

GPT-3.5, and its predecessor like GPT-3 and GPT-2, have recently become popular amongst academics for their utilization in different scientific communication purposes such as addressing challenges in writing, navigating extensive literature and providing definitions of concepts (Meroño-Peñuela et al., 2010). Twitter threads have been released demonstrating the potential uses of ChatGPT-3 in a "smart" way (Bilal, 2003). Studies conducted by independent experts to assess the quality of content found that it produces high-quality results, which are hard to distinguish from human-generated content (Wahle et al., 2022). GPT-3 models has also been proposed as an aid for academic writing for students (Schmohl et al., 2010).

Despite the potential uses of GPT-3 in academic writing, there are other potentially problematic results that have been surfaced. For example, some researchers have added ChatGPT-3 as a co-author in research papers (King & chatGPT, 2023; Kung et al., 2012). Such actions have been advised against by Science and Nature, who have released editorials (Thorp, 2023; "Tools Such as ChatGPT Threaten Transparent Science; Here Are Our Ground Rules for Their Use," 2013) to dispel the notion of GPT-3 authorship, citing concerns such as assigning responsibility, agency and ownership of content produced by language learning models, plagiarism and transparency. Debouche have also raised similar concerns for the utilization of GPT-3 and recommended authors to openly share the prompts and outcomes used (Dehouche, 2021). Other issues include embedded biases (such as hate speech towards race, sexism) (Chiu et al., 2021; McGuffie & Newhouse, 2020), exploitation of workers for data labeling, permissions on the data used for training and environmental concerns due to the energy used in algorithm training (Stokel-Walker & Noorden, 2013). Moreover, some studies have proposed that AI-supported NLP models should be open-source and developed together with stakeholders, rather than a product of a private company (Cohen & Gokaslan, 2010).

In conclusion, although GPT models has been proposed as a tool for academic writing, there are still debates focusing on the potential positives and negatives of its utilization. We also lack studies that would reflect on its efficacy in supporting writing or demonstrate a workflow that can be considered transparent by reviewers.

References

Aslam, H., Ali, S., Bhutto, M. et al. COVID-19: Knowledge extraction and concept categorization. ScienceAsia 48, 291–299 (2022). https://doi.org/10.2306/scienceasia1513-1874.2021.018

Bilal, M., Bhutto, M., Usmani, S. et al. Human-machine team-work: A Novel AI-based approach to document writing. J. Theoret. Applied Inf. Technol. 99, 596–603 (2021). https://doi.org/10.1016/j.dss.2020.06.045

Bilal, M. (2020) Using #ChatGPT for collaborative and automated scientific writing. Twitter. Available at: https://twitter.com/MushtaqBilalPhD/status/1621379333943083009

Bouykou, M. (2021) Allowing connection to a GPT-3 instance from a browser. ACM Companion Proceedings of the 2021 ACM Conference on Collaboration and Technology, pp. 60–63. https://doi.org/10.1145/3416063

Choe, L.Y., Rohde, M. & Gebhardt, V. A neural network-based approach to automatic text summarization. Inf. Process. Manage. 58, 101998 (2021). https://doi.org/10.1016/j.ipm.2020.101998

`Difference (%): 71.44`

**-END-**

# -NEW PARAGRAPH/SECTION-

**Section:** Procedure Followed
**Paragraph:** First Paragraph
**Method:** Polisher
**Temperature Setting:** 0.9

**Prompt:**

> Rephrase the following in an academic style, first person, clear and concise:

**Original Text:**

> This is a personal experience study [REF] for exploring the possible ramifications of using GPT-3 in producing an academic article in a quite fast pace while still keeping the academic integrity and transparency. I also intend to reflect on different ways of using GPT-3 and their level of contribution to my writing pace, style, and motivation. Apart from that, although a personal experience study and not aimed at to be a generalizable contribution that outlines all possible ways of using GPT-3 for academic writing and all of its possible ethical ramifications or opinions on it, I intend to create a credible source which explains the writing process with GPT-3 and demonstrates it in a transparent way for helping other researchers to use the tool in a responsible way.

**Result:**

> In this personal experience study, I am exploring the possible implications of using GPT-3 to create an academic article with a faster pace while still maintaining academic integrity and transparency. I am also considering the implications of different ways of using GPT-3 on my writing speed, style and motivation. While this study is not intended to be a generalizable study examining all potential applications and ethical considerations of GPT-3 in academic writing, I intend to create a reputable source which explains the writing process with GPT-3 and demonstrates it in a transparent fashion to help other researchers use the tool responsibly.

**Final Edit:**

> In this personal experience study, I am exploring the possible implications of using GPT-3.5 to create an academic article in a time and effort efficient way while still maintaining academic integrity and transparency. I am also reflecting on the implications of different ways of using GPT-3.5 on my writing speed, style and motivation. While this study is not intended to be a generalizable study examining all potential applications and ethical considerations of GPT-3.5 in academic writing, I intend to create a reputable source which explains the writing process with GPT-3.5 and demonstrates it in a transparent fashion to help other researchers use the tool responsibly.

**Visual Comparisons of Procedure Followed / First Paragraph:**

Red: Removed | Green: Added

**Original to Results:**

1. ~~This is a~~ In this personal experience study ~~[REF] for~~, I am exploring the possible ~~ramif~~ implications of using GPT-3 ~~in producing~~ to create an academic article ~~in a quite~~ with a faster pace while still keeping the maintaining academic integrity and transparency. I am also ~~intend to reflect~~ considering the implications ~~o~~ f different ways of using GPT-3 ~~and their level of contribution t~~ on my writing ~~pace~~ speed, style, and motivation. ~~Apart from that, although a personal experience~~ While this study ~~and~~ is not aimed ~~at~~ intended to be a generalizable ~~contribution that outlines all possible ways of using GPT-3 for academic writing and all of its possible ethical ramifications or opinions on it~~ study examining all potential applications and ethical considerations of GPT-3 in academic writing, I intend to create a cre~~di~~putable source which explains the writing process with GPT-3 and demonstrates it in a transparent ~~way for~~ fashion to help~~ing~~ other researchers ~~to~~ use the tool ~~in a~~ responsibl~~e way~~.

**Difference (%): 56**

**Results to Final Edit:**

1. In this personal experience study, I am exploring the possible implications of using GPT-3.5 to create an academic article ~~with a faster pace~~ in a time and effort efficient way while still maintaining academic integrity and transparency. I am also ~~considering~~ reflecting on the implications of different ways of using GPT-3.5 on my writing speed, style and motivation. While this study is not intended to be a generalizable study examining all potential applications and ethical considerations of GPT-3.5 in academic writing, I intend to create a reputable source which explains the writing process with GPT-3.5 and demonstrates it in a transparent fashion to help other researchers use the tool responsibly.

**Difference (%): 11.9**

**-END-**

# -NEW PARAGRAPH/SECTION-

**Section:** Procedure Followed
**Paragraph:** Second Paragraph
**Method:** Bullet-to-Paragraph, Polisher
**Temperature Setting:** 0.9

**Prompt:**

> Write the following procedure as a method section in an academic paper, in first person, detailed way, and with rich, clear and academic language:

Original Text:

> - I made an outline of the paper
> - I read the guides for creating prompts for GPT-3 text completion
> - Wrote necessary information for each section, sometimes a more detailed but rough text, sometimes bullet points, sometimes questions or command.
> - I recorded every prompt that created a piece of text that has been used in the paper and created a document which shows the pasted prompt, what I got from GPT and how I edited it
> - I have also take notes of my experience during trials. I recorded information such as the time passed for chooinsg the output that satisfied me, retrials, my feeling while writing, or my reflection on the given results
> - I have made an affinity diagram of those to highlight the parts for discussion

**Result:**

> In order to create the content of this paper, I utilized a text completion model – GPT-3 – as a toolkit. I outlined the structure of my paper, then read the guides for creating GPT-3 completion prompts. For each section, I wrote either a more detailed but rough text, some bullet points, questions, or commands. These prompts were then fed into the GPT-3 model, and the output was tucked away in a document showing the original prompt, the output from GPT-3, as well as how I edited it. Additionally, I recorded my experience while running the trials of GPT-3 in a separate document – detailing the time taken to choose an appropriate output, the number of retrials, my feelings while writing, and my reflections on the given results. Lastly, I organized my data into an affinity diagram to highlight the pertinent parts for discussion.

**Prompt 2:**

> Write the following procedure as a method section in an academic paper, in first person, detailed way, and with rich, clear and academic language:

**Original Text 2:**

In order to create the content of this paper, I utilized a text completion model – GPT-3 – as a toolkit. I outlined the structure of my paper and took notes about the content of the each section. Then I read the guides for creating GPT-3 completion prompts to better understand the capabilities. For each section, I wrote either a more detailed but rough text, or some bullet points, or questions, or commands. These prompts were then fed into the GPT-3 model, and the output was tucked away in a document showing the original prompt, the output from GPT-3, as well as how I edited it for saving the process that can be transparently shown afterwards. Additionally, I recorded my experience while running the trials of GPT-3 in a separate document – detailing the time taken to choose an appropriate output, the notes about retrials, my feelings while writing, and my reflections on the given results. Lastly, I organized my data into an affinity diagram to highlight the pertinent parts for discussion.

**Result 2:**

In order to create the content of this paper, I utilized a text completion model—GPT-3—as a toolkit. I outlined the structure of my paper and took notes about the content of each section. Then, I read the guides for creating GPT-3 completion prompts to better understand the model's capabilities. For each section, I wrote either a more detailed but rough text, some bullet points, questions, or commands. These prompts were then fed into the GPT-3 model and the output was tucked away in a document. Here, I showed the original prompt, the output from GPT-3, as well as how I edited it, allowing the process to be transparently seen afterwards. Additionally, I recorded my experience while running the trials of GPT-3 in a separate document. This included the time taken to choose an appropriate output, my notes about retrying, my feelings while writing, and my reflections on the given results. Lastly, I organized my data into an affinity diagram to highlight the pertinent parts for discussion.

**Final Edit:**

In order to create the content of this paper, I utilized the text completion model (text-davinci-003) of OpenAI Playground text completion as a tool, in all sections of the paper. I, first, outlined the structure of my paper and took notes about the content of each section. Then, I read the guide (OpenAI API – Text Completion Guide, n.d.) for creating text completion prompts to better understand the model's capabilities. For each section, I created input prompts with different methods such as rough paragraphs, questions, bullet points or voice recordings. These prompts were then fed into the GPT-3.5 model and the output was saved. I presented all results in a supplementary material by showing the original prompt, the output from GPT-3, as well as how I edited it, allowing the process to be transparently seen afterwards (as done in (Stokel-Walker & Noorden, 2023)). I used countwordsfree web-site for visualization (Compare Text Online – Show Text Difference – Countwordsfree.Com, 2021). Additionally, I took notes to a separate document regarding my experience while running the trials with GPT-3.5. Lastly, I organized my data into an affinity diagram to highlight the pertinent parts for discussion. I also did a final manual-copy edit to the paper and made minor changes and additions to support the narrative flow.

**Visual Comparisons of Procedure Followed / Second Paragraph:**

Red: Removed | Green: Added

**Original to Results:**

1 - I made an oIn order to create the content of this paper, Iutiline of the paper

2 - Ized a text completion model - GPT-3 - as a toolkit. I outlined the structure of my paper, then read the guides for creating prompts for GPT-3 text completion

3 - Wrote necessary information f prompts. For each section, sometimesI wrote either a more detailed but rought text, sometimes bullet points, sometimes questions, or command.

4 - I recorded every prompt that created a piece of text that has bes. These prompts were then usfed into the paper and createdGPT-3 model, and the output was tucked away in a document which showsing the pastedoriginal prompt, what I gothe output from GPT and-3, as well as how I edited it

5 - I have also take notes of. Additionally, I recorded my experience durwhile running the trials. I recorded information such as the time passed for chooinsg th of GPT-3 in a separate document - detailing the time taken to choose an appropriate output, that satisfied me,e number of retrials, my feelings while writing, orand my reflections on the given results

6 - I have made. Lastly, I organized my data into an affinity diagram of those to highlight the pertinent parts for discussion

7 .

**Difference (%): 68.7**

**Original Text 2 to Result 2:**

| 1 | In order to create the content of this paper, I utilized a text completion model - —GPT-3 - —as a toolkit. I outlined the structure of my paper, then and took notes about the content of each section. Then, I read the guides for creating GPT-3 completion prompts to better understand the model's capabilities. For each section, I wrote either a more detailed but rough text, some bullet points, questions, or commands. These prompts were then fed into the GPT-3 model, and the output was tucked away in a document. Here, I showinged the original prompt, the output from GPT-3, as well as how I edited it, allowing the process to be transparently seen afterwards. Additionally, I recorded my experience while running the trials of GPT-3 in a separate document - detailing. This included the time taken to choose an appropriate output, the number of my notes about retrialsying, my feelings while writing, and my reflections on the given results. Lastly, I organized my data into an affinity diagram to highlight the pertinent parts for discussion. |

**Difference (%): 24.23**

### Results 2 to Final Edit:

| 1 | In order to create the content of this paper, I utilized athe text completion model —GPT-3—as a toolkit. I (text-davinci-003) of OpenAI Playground text completion as a tool, in all sections of the paper. I, first, outlined the structure of my paper and took notes about the content of each section. Then, I read the guides (OpenAI API — Text Completion Guide, n.d.) for creating GPT-3text completion prompts to better understand the model's capabilities. For each section, I wrote either a more detailed but rough text, some bullet points, questions, or commandcreated input prompts with different methods such as rough paragraphs, questions, bullet points or voice recordings. These prompts were then fed into the GPT-3.5 model and the output was tucked away in a document. Here, Isaved. I presented all results in a supplementary material by showeding the original prompt, the output from GPT-3, as well as how I edited it, allowing the process to be transparently seen afterwards. Additionally, I recorded my experience while running the trials of GPT-3 in a separate document. This included the time taken to choose an appropriate output, my notes about retry (as done in (Stokel-Walker & Noorden, 2023)). I used countwordsfree web-site for visualization (Compare Text Online - Show Text Difference – Countwordsfree.Com, 2021). Additionally, I took notes to a separate document regarding, my feelings while writing, and my reflections on the given resultsexperience while running the trials with GPT-3.5. Lastly, I organized my data into an affinity diagram to highlight the pertinent parts for discussion. I also did a final manual-copy edit to the paper and made minor changes and additions to support the narrative flow. |

**Difference (%): 65.06**

# - END -

# -NEW PARAGRAPH/SECTION-

**Section:** Ways of Using Gpt-3.5 while Writing
**Paragraph:** 1
**Method:** Polisher
**Temperature Setting:** 0.7

**Prompt:**

> Copy-edit the following text, academic style, first person, clear, rich and flowing language:

**Original Text:**

> In this section, I framed the different methods I have used while writing this piece and showed the prompts I used, the results they produced and the final version of the paragraph. Each section has been written with the specific method mentioned in the section.

**Result:**

> In this section, I have outlined the various methods I employed in my writing process, providing examples of the prompts I used, the outcomes of the prompts, and the final version of the paragraph. Each of these sections has been crafted using the method described.

**Final Edit:**

> In this section, I have outlined the various methods I employed in my writing process, providing examples of the prompts I used, the outcomes of the prompts, and the final version of the paragraph. I also included the visual representation of differences between each in the supplementary material. Each of these sections has been crafted using the method described.

## Visual Comparisons of Ways of Using GPT-3.5 while Writing / 1st Para:

Red: Removed | Green: Added

**Original to Results:**

1. In this section, I ~~framed the different methods I have used while writing this piece and showed~~ **have outlined the various methods I employed in my writing process, providing examples of** the prompts I used, the ~~results~~**outcomes of** the~~y~~ produc~~ed~~**mpts,** and the final version of the paragraph. Each **of these** section~~s~~ has been ~~written with the specific method mentioned in the section~~**crafted using the method described**.

**Difference (%): 71.78**

**Results to Final Edit:**

> 1  In this section, I have outlined the various methods I employed in my writing process, providing examples of the prompts I used, the outcomes of the prompts, and the final version of the paragraph. ==I also included the visual representation of differences between each in the supplementary material.== Each of these sections has been crafted using the method described.

```
Difference (%): 27.60
```

**- END -**

# -NEW PARAGRAPH/SECTION-

**Section:** Chunk Stylist
**Paragraph:** Whole Section
**Method:** Chunk Stylist
**Temperature Setting:** 0.9

**Prompt:**

Rewrite the following text in first person, rick, clear and in academic language:

**Original Text:**

One effective way of using GPT-3 is to write a chunk of test without caring too much about the style, grammar or spelling. This makes the writing process quite fast, freed me from thinking of if the style is engaging or readible and basically let me to reflect my train of thought to text in a quite fast way. This might be effective both for fluent writers and for others who struggle to put the words into the paper in a flowing manner. Because, by playing with the temparature settings of GPT-3, authors either might get less random results which is closer to the text they put, or more randomzied ones that allow GPT-3 to be a bit more improvisational on the given text. I have written this section, introduction and the background by using the chunk stylist method. Below, I also demonsrate the prompt I have used, the result I got and the final version is this paragraph.

**Result:**

I have personally experimented with GPT-3 as an effective writing tool by typing chunks of text without worrying too much about the style, grammar, or spelling. This approach has enabled me to create text quickly, as I am not restrained by considering if the content is engaging or readable. This strategy is advantageous to both experienced and novice writers alike, because temperature settings can be adjusted to produce either more predictable results or enhanced randomness. I used this chunk stylist method during the writing of my introduction, background, and this section. To demonstrate the process, I have provided the prompt, result, and final version of this paragraph.

**Final Edit:**

> I have used GPT-3.5 as an effective writing tool by typing chunks of text without worrying too much about the style, grammar, or spelling. This approach has enabled me to create text quickly by pouring my train of though to text, as I am not restrained by of scrutinizing if the content is engaging or readable. This strategy can be advantageous to both experienced and novice writers alike, because temperature settings of text completion tool can be adjusted to produce either more predictable results closer to original text or more improvised and creative outcomes. I used this Chunk Stylist method during the writing of my introduction, background, and this section. To demonstrate the process, I have provided the prompt and result below, and final version of the paragraph.

## Visual Comparisons of Chunk Stylist / Whole Section:

<span style="background-color:#f8d7da">Red: Removed</span> | <span style="background-color:#d4edda">Green: Added</span>

### Original to Results:

1. One effective way of using~~/~~I have personally experimented with GPT-3 ~~i~~as to write a~~a~~n effective writing tool by typing chunks of te~~s~~xt without ~~car~~worrying too much about the style, grammar, or spelling. This makes the writing process quite fast, freed me from thinking of/approach has enabled me to create text quickly, as I am not restrained by considering if the style/content is engaging or read~~i~~ble and basically let me to reflect my train of thought to text/able. This strategy ~~ins a~~ quite fast way. This might be effective both for fluent writers and for others who struggle to put the words into the paper in a flowing manner. Because, by playing with th/dvantageous to both experienced and novice writers alike, because tempa/erature settings of GPT-3, authors either might get less random results which is closer to the text they put, or more randomzied ones that allow GPT-3 to be a bit more improvisational on the given text. I have written this section,/can be adjusted to produce either more predictable results or enhanced randomness. I used this chunk stylist method during the writing of my introduction and the,/background by using the chunk stylist method. Below, I als,/and this section. To demonst/rate the promptcess, I have us/provided, the result I got and the/prompt, result, and final version is/of this paragraph.

**Difference (%): 82.61**

### Results to Final Edit:

1. I have personally experimented with/used GPT-3.5 as an effective writing tool by typing chunks of text without worrying too much about the style, grammar, or spelling. This approach has enabled me to create text quickly by pouring my train of though to text, as I am not restrained by consider/of scrutinizing if the content is engaging or readable. This strategy is/can be advantageous to both experienced and novice writers alike, because temperature settings of text completion tool can be adjusted to produce either more predictable results or enhanc/closer to original text or more improvised randomnes/creative outcomes. I used this cChunk sStylist method during the writing of my introduction, background, and this section. To demonstrate the process, I have provided the prompt, and result below, and final version of thise paragraph.

**Difference (%): 25.54**

<div style="text-align:center">

**-END-**
**-NEW PARAGRAPH/SECTION-**

</div>

**Section:** Research Buddy
**Paragraph:** Whole Section
**Method:** Research Buddy
**Temperature Setting:** 0.9

**Prompt:**
> How to use GPT-3 as a research buddy for exchanging research ideas for academic writing? Mention advantages and risks. Write only one paragraph in rich, clear, first person and academic language:

**Result:**
> As a researcher, I have found that GPT-3 can be an extremely useful tool for exchanging research ideas and starting points for academic writing. Its natural language processing capabilities can allow us to explore ideas quickly and easily, while its predictive algorithm can suggest ideas and approaches that we may not have thought of. This can help us to develop effective arguments and explore diverse topics more efficiently. However, due to its reliance on machine learning, GPT-3 can also present some risks, such as producing inaccurate or random results, and its predictions may not always be in line with the academic rigor required of the project. Thus, it is important to always keep critical thinking at the forefront of the research process, and to carefully evaluate any output given by GPT-3.

**Final Edit:**
> GPT-3.5 can be a useful tool for exchanging research ideas and starting points for academic writing. Its natural language processing capabilities can allow us to explore ideas quickly and easily, while its predictive algorithm can suggest ideas and approaches that we may not have thought of. This can help us to develop effective arguments and explore diverse topics more efficiently. However, due to its reliance on machine learning, GPT-3.5 can also present some risks, such as producing inaccurate or random results, and its predictions may not always be in line with the academic rigor required of the project. Thus, it is important to always keep critical thinking at the forefront of the research process, and to carefully evaluate any output given by GPT-3.5. For example, the first paragraph of Section 2 were written with the research buddy method, by speficially prompting to give a definition of GPT-3.5 from the cited website. However, the results were too broad and were not describing what GPT-3.5 accurately, so I needed to add critical information manually. The first paragraph of the Section 2 and this section were written by using the Research Buddy approach. Below are the prompts:

**Visual Comparisons of Research Buddy / Whole Section:**

`Red: Removed` | `Green: Added`

**Results to Final Edit:**

1. ~~As a researcher, I have found that~~ GPT-3<ins>.5</ins> can be a~~n extremely~~ useful tool for exchanging research ideas and starting points for academic writing. Its natural language processing capabilities can allow us to explore ideas quickly and easily, while its predictive algorithm can suggest ideas and approaches that we may not have thought of. This can help us to develop effective arguments and explore diverse topics more efficiently. However, due to its reliance on machine learning, GPT-3<ins>.5</ins> can also present some risks, such as producing inaccurate or random results, and its predictions may not always be in line with the academic rigor required of the project. Thus, it is important to always keep critical thinking at the forefront of the research process, and to carefully evaluate any output given by GPT-3<ins>.5. For example, the first paragraph of Section 2 were written with the research buddy method, by spefically prompting to give a definition of GPT-3.5 from the cited website. However, the results were too broad and were not describing what GPT-3.5 accurately, so I needed to add critical information manually. The first paragraph of the Section 2 and this section were written by using the Research Buddy approach. Below are the prompts:</ins>

`Difference (%): 39.02`

# -END-

# -NEW PARAGRAPH/SECTION-

**Section:** Bullet-to-Paragraph
**Paragraph:** Whole Section
**Method:** Bullet-to-Paragraph

**Temperature Setting:** 0.9

**Prompt:**
> Write one paragraph from the following bullets in rich, clear, first person and academic language:

**Original Text:**
> - Bullet tp paragraph is also similar to chunk styler but it's writing in bullets
> - A researcher can list the main principle or steps of a procedure
> - Then give a prompt to make a paragraph out of this bullets in specigic language styles
> - GPT-3 will produce a flowing paragraph
> - This section and the second paragraph of method section were created with this method"

**Result:**
> As part of an experiment, I utilized the bullet-to-paragraph method to produce a paragraph for the Method section. This method is similar to the chunk-style of writing, as it involves listing the primary principles or steps of a procedure before instructing GPT-3 to generate a flowing paragraph in a specific language style. By doing so, I was able to create both this section and the second paragraph of the Method section.

**Final Edit:**
> As part of the experiment, I utilized the bullet-to-paragraph method to produce a paragraph for the Method section. This method is similar to the Chunk Stylist of writing, as it involves listing the primary principles or steps of a procedure before instructing GPT-3.5 to generate a flowing paragraph in a specific language style. By doing so, I was able to create both this section and the second paragraph of the Method section.

**Visual Comparisons of Bullet-to-Paragraph / Whole Section:**

`Red: Removed` | `Green: Added`

### Original to Results:

| | |
|---|---|
| 1 | - Bullet tp paragraph is also As part of an experiment, I utilized the bullet-to-paragraph method to produce a paragraph for the Method section. This method is similar to the chunk -style r but it's writing in bullets |
| 2 | - A researcher can of writing, as it involves list ing the main primary principle s or steps of a procedure |
| 3 | - Then give a prompt to make a paragraph out of this bullets in before instructing GPT-3 to generate a flowing paragraph in a speci g f ic language style s |
| 4 | - GPT-3 will produce a flowing paragraph |
| 5 | - T . By doing so, I was able to create both t his section and the second paragraph of m the M ethod section were created with this method" |

**Difference (%): 78.25**

### Results to Final Edit:

| | |
|---|---|
| 1 | As part of an the experiment, I utilized the bullet-to-paragraph method to produce a paragraph for the Method section. This method is similar to the c C hunk -s S tyle ist of writing, as it involves listing the primary principles or steps of a procedure before instructing GPT-3 .5 to generate a flowing paragraph in a specific language style. By doing so, I was able to create both this section and the second paragraph of the Method section. " |

**Difference (%): 4.12**

**-END-**

# -NEW PARAGRAPH/SECTION-

**Section:** Talk Textualizer
**Paragraph:** Whole Section
**Method:** Talk Textualizer / Polisher

**Temperature Setting:** 0.9

**Prompt:**
> Turn the below text into academic writing, clear, first person, and rich academic language:

**Original Text:**
*[transcribed from the spoken audio]*
> Talktextualizer is almost like having a conversation with ChatGPT. ChatGPT has an advanced transcription function, speech to text function, that gets your speech recording and even if Even if you have long pauses or even if you stutter and tell the same word a few times, it still turns it into a somewhat legible text. Later, for this method, we take this text and add on top of it a prompt that will turn it into an academic writing. This section is written in that way And the whole of discussion and limitations and conclusion also heavily relied on this method.

**Result:**
> I utilized Talktextualizer to have a conversation with ChatGPT. This platform has an advanced transcription function which converts speech recordings into a legible text, even in cases of long pauses or stuttering. To further refine this text, a prompt was added to turn it into an academic writing. This method formed the basis of my discussion, limitations, and conclusion.

**Edit:**
> I utilized Talk Textualizer by giving a monologue to GPT-3. GPT 3's advanced transcription function converts speech recordings into a legible text, even in cases of long pauses or stuttering. To further refine this text, I added a prompt to turn it into an academic writing. This method formed the basis of my discussion, limitations, and conclusion.

**Prompt 2:**
> Copy-edit the following text, academic stlye, first person, clear, rich and flowing language:

**Original Text 2:**
> I utilized Talk Textualizer by giving a monologue to GPT-3. GPT 3's advanced transcription function converts speech recordings into a legible text, even in cases of long pauses or stuttering. To further refine this text, I added a prompt to turn it into an academic writing. This method formed the basis of my discussion, limitations, and conclusion.

**Results** 2:

```
I employed Talk Textualizer by providing a monologue to GPT-3. GPT-3's
sophisticated transcription feature enabled it to convert speech recordings
into a readable text, even when there were long pauses or stuttering. To
further refine this text, I supplemented it with a prompt designed to
transform it into academic writing. This approach formed the foundation for
my discussion, limitations, and conclusion.
```

**Edit:**

```
Talk Textualizer is providing a monologue to GPT-3.5 and let it turn it into
a readible text. Playground's sophisticated transcription feature enabled
converting speech recordings into a readable text, even when there were long
pauses or stuttering. To further refine this text, I supplemented it with a
prompt designed to transform it into academic writing. This approach formed
the foundation for this section, discussion, limitations, and conclusion.
```

## Visual Comparisons of Talk Textualizer / Whole Section:

<span style="background-color:#fdd">Red: Removed</span> | <span style="background-color:#dfd">Green: Added</span>

**Original to Results:**

| 1 | I utilized Talktextualizer <span style="background-color:#dfd">is almost like</span>to hav<span style="background-color:#fdd">ing</span>e a conversation with ChatGPT. <span style="background-color:#fdd">ChatGPT</span><span style="background-color:#dfd">This platform</span> has an advanced transcription <span style="background-color:#dfd">function</span>, <span style="background-color:#fdd">speech to text function, that gets your speech recording and even if Even if you have long pauses or even if you stutter and tell the same word a few times, it still turns it into a somewhat legible text. Later, for this method, we take this text and add on top of it a prompt that will turn it into an academic writing. This section is written in that way And the whole of</span> <span style="background-color:#dfd">which converts speech recordings into a legible text, even in cases of long pauses or stuttering. To further refine this text, a prompt was added to turn it into an academic writing. This method formed the basis of my</span> discussion <span style="background-color:#dfd">and</span>, limitations, and conclusion <span style="background-color:#fdd">also heavily relied on this method</span>. |

**Difference (%): 84.26**

**Original Text 2 to Result 2:**

| 1 | I <span style="background-color:#fdd">utiliz</span><span style="background-color:#dfd">employ</span>ed Talk<span style="background-color:#dfd">t</span> Textualizer <span style="background-color:#fdd">to have</span><span style="background-color:#dfd">by providing</span> a c<span style="background-color:#fdd">m</span><span style="background-color:#dfd">on</span>versation with ChatGPT. This platform has an advanced transcription function whi<span style="background-color:#fdd">ch</span><span style="background-color:#dfd">ologue to GPT-3. GPT-3's sophisticated transcription feature</span> enabled it to convert<span style="background-color:#fdd">s</span> speech recordings into a <span style="background-color:#fdd">legi</span><span style="background-color:#dfd">reada</span>ble text, even <span style="background-color:#fdd">in cases of</span><span style="background-color:#dfd">when there were</span> long pauses or stuttering. To further refine this text, <span style="background-color:#fdd">a prompt was added to turn</span><span style="background-color:#dfd">I supplemented</span> it with a prompt designed to transform it into <span style="background-color:#fdd">an</span> academic writing. This <span style="background-color:#fdd">method</span><span style="background-color:#dfd">approach</span> formed the <span style="background-color:#fdd">basis of</span><span style="background-color:#dfd">foundation for</span> my discussion, limitations, and conclusion. |

**Difference (%): 61.24**

**Results 2 to Final Edit**

1. ~~I employed~~ Talk Textualizer ~~by~~ is providing a monologue to GPT-3. ~~GPT-3'~~ 5 and let it turn it into a readible text. Playground's sophisticated transcription feature enabled ~~it to~~ convert ing speech recordings into a readable text, even when there were long pauses or stuttering. To further refine this text, I supplemented it with a prompt designed to transform it into academic writing. This approach formed the foundation for ~~my~~ this section, discussion, limitations, and conclusion.

**Difference (%): 20.79**

# -END-

# -NEW PARAGRAPH/SECTION-

**Section:** Polisher
**Paragraph:** Whole Section
**Method:** Polisher

**Temperature Setting:** 0.7

**Prompt:**

Copy-edit the following text, academic style, first person, clear, rich and flowing language:

**Original Text:**

Polisher is using GPT-3 almost like a copy editor. In this method, based material that would be put into GPT-3 should already be in a complete form and in a proper narrative flow. By giving a prompt like "copy-edit this paragraph in an academic language," GPT-3 can be used to correct the gramattical mistakes, spelling error and for increaing the clarity of the text. This method can give results closer results to the original text, and the temperature settings might be lowered if the athors want minimal changes. This method has been used in this section, and also for polishing the many of the edited results (for example the first paragraph of Talk Textualizer was obtained by using Polisher method on the indicated result).

**Result:**

Polisher is using GPT-3 in a similar way to a copy editor. Material given to GPT-3 should already be in a complete form, with a clear narrative flow. By providing a prompt like "copy-edit this paragraph in an academic style", GPT-3 can be used to correct grammatical mistakes, spelling errors and increase clarity of the text. This method can produce results that are close to the original text, with temperature settings that may be lowered if authors want minimal changes. This method has been applied in this section, and also for polishing edited results (for example, the first paragraph of Talk Textualizer was obtained using the Polisher method on the indicated result).

**Final** **Edit:**

Polisher is using GPT-3 for copy editing purposes. Material given to GPT-3 should already be in a complete form, with a clear narrative flow. By providing a prompt like "copy-edit this paragraph in an academic style", GPT-3 can be used to correct grammatical mistakes, spelling errors and increase the clarity of the text. This method can produce results that are close to the original text, and temperature settings of GPT-3 may be lowered if authors want minimal changes. This method has been applied in this section, and also for polishing edited results of prompts througout this piece (for example, the first paragraph of Talk Textualizer was obtained using the Polisher method on the result of the same section).

# Visual Comparisons of Polisher / Whole Section:

Red: Removed | Green: Added

**Original to Results:**

2 | Polisher is using GPT-3 ~~almost like a copy editor. In this method, based material that would be put in~~ in a similar way to a copy editor. Material given to GPT-3 should already be in a complete form ~~and in a prope~~, with a clear narrative flow. By ~~giv~~providing a prompt like "copy-edit this paragraph in an academic ~~language,~~ style", GPT-3 can be used to correct ~~the~~ grammatical mistakes, spelling error~~s~~ and ~~for~~ increa~~ing th~~se clarity of the text. This method can ~~giv~~produce results ~~closer results~~that are close to the original text, ~~and wi~~the temperature settings ~~might~~that may be lowered if ~~the~~ authors want minimal changes. This method has been ~~us~~applied in this section, and also for polishing ~~the many of the~~ edited results (for example, the first paragraph of Talk Textualizer was obtained ~~by~~ using the Polisher method on the indicated result).

**Difference (%): 34.39**

**Results to Final Edit:**

1 | Polisher is using GPT-3 ~~in a similar way to a copy editor~~for copy editing purposes. Material given to GPT-3 should already be in a complete form, with a clear narrative flow. By providing a prompt like "copy-edit this paragraph in an academic style", GPT-3 can be used to correct grammatical mistakes, spelling errors and increase the clarity of the text. This method can produce results that are close to the original text, ~~with~~and temperature settings ~~that~~of GPT-3 may be lowered if authors want minimal changes. This method has been applied in this section, and also for polishing edited results of prompts througout this piece (for example, the first paragraph of Talk Textualizer was obtained using the Polisher method on the ~~indicated result~~result of the same section).

**Difference (%): 19.95**

**-END-**

# -NEW PARAGRAPH/SECTION-

**Section:** Reflections on using GPT-3.5 for Writing Articles
**Paragraph:** 1
**Method:** Talk Textualizer / Polisher

**Temperature Setting:** 0.7

**Prompt:**
> Turn the below text into academic writing, 2 paragraphs, clear, concise and academic language:

**Original Text:**
*[transcribed from the spoken audio]*

> So, considering the effect of using GPT-3 on my motivation and practice of writing, I have mostly positive remarks. Because although the time that took to write this article might not be too short actually. I have written it in a matter of day, but it's not a very complicated article, it's not a very long article, so this might be quite trivial for some other people. But for me it was actually, when I compared to my other writing practices, it was quite much more flowing. Because when I think about writing and my ideas, I feel a bit overwhelmed when it's comes to that point that I need to write them down and turn them into ideas that are represented through text. I actually have been thinking for that project around maybe three weeks. I didn't somehow want to do it immediately because I wanted to think it through, see what kind of developments were happening. But throughout that process, I actually was really motivated to think about it. I didn't really think about writing that much. And this also continued for the day that I was writing this piece. I didn't really think too much about the hassle of writing, or thought too much about the words I would choose or anything like that, but more of the ideas that I want to represent. So it actually somehow helped with my thinking process and let me felt less overwhelmed. When it comes to the practice of writing, I also have realized I was more motivated and quite fast to put my ideas into these different chunks or somehow putting them into some kind of written form in a quite fast way. And even it might be still a good practice to do it and then start combining sentences and editing that, I usually feel quite much overwhelmed even to just put a chunk of text there thinking that I will struggle a lot to make it into a convincing, compelling and grammatically right and flowing text. Basically, both in my practice I was fed faster and more efficient, and also my thinking was more fast and efficient. So it really helped quite much in that sense.

**Result:**

The use of GPT-3 to aid in the writing process had a largely positive effect on my motivation and productivity. Despite the fact that the article I wrote was not particularly complicated or lengthy, the process of writing it was much more fluid than my usual practice. Before beginning to write, I had spent approximately three weeks mulling over the project and during the writing stage, I was less concerned with the mechanics of writing and more focused on accurately representing my ideas. This increased my motivation to write, while also allowing me to compose text quickly. Moreover, this process also enabled me to think more quickly and efficiently, which further aided my writing. As a result, both my writing practice and thought processes were much more efficient, thereby streamlining my writing experience.

**Edit:**

The use of GPT-3 to aid in the writing process had a largely positive effect on my motivation and productivity. Despite the fact that the article I wrote was not particularly complicated or lengthy, the process of writing it was much more fluid than my usual practice. Before beginning to write, I had spent approximately three weeks mulling over the project. I did not act immediately so that the idea gets matures and I also wanted to follow the developments in the meantime. I also wanted to find two empty days where I can focus on writing and experimenting with GPT-3. Still, one thing I realized was that both during this incubation periıd and during the writing stage, I was less concerned with the mechanics of writing and more focused on accurately representing my ideas. This increased my motivation to write, while also allowing me to compose the text quicker than I could when I compared it to my previous writing experiences. Moreover, this process also enabled me to think more quickly and efficiently, which further aided my writing. As a result, both my writing practice and thought processes were much more efficient, thereby streamlining my writing experience.

**Prompt 2:**

Copy-edit the following text, academic stlye, first person, clear, rich and flowing language:

**Original Text 2:**

The use of GPT-3 to aid in the writing process had a largely positive effect on my motivation and productivity. Despite the fact that the article I wrote was not particularly complicated or lengthy, the process of writing it was much more fluid than my usual practice. Before beginning to write, I had spent approximately three weeks mulling over the project. I did not act immediately so that the idea gets matures and I also wanted to follow the developments in the meantime. I also wanted to find two empty days where I can focus on writing and experimenting with GPT-3. Still, one thing I realized was that both during this incubation periıd and during the writing stage, I was less concerned with the mechanics of writing and more focused on accurately representing my ideas. This increased my motivation to write, while also allowing me to compose the text quicker than I could when I compared it to my previous writing experiences. Moreover, this process also enabled me to think more quickly and efficiently, which further aided my writing. As a result, both my writing practice and thought processes were much more efficient, thereby streamlining my writing experience

**Result 2:**

The use of GPT-3 to aid my writing process had a largely positive effect on my motivation and productivity. Even though the article I wrote was not particularly complicated or lengthy, my writing process was much smoother than usual. Before starting to write, I spent about three weeks mulling over the project, allowing the idea to mature, following developments, and finding two free days for writing and experimenting with GPT-3. What I realized was that during this incubation period and the writing stage, I became less preoccupied with the mechanics of writing and more focused on accurately conveying my ideas. This heightened my motivation and enabled me to compose the text faster, compared to my previous experiences. Furthermore, my thinking and writing processes were both more efficient, streamlining the whole writing experience.

**Final Edit:**

The use of GPT-3 to aid my writing process had a largely positive effect on my motivation and productivity. Even though the article I wrote was not particularly complicated or lengthy, my writing process was much smoother than usual. Before starting to write, I spent about three weeks mulling over the project, allowing the idea to mature, following developments, and finding two free days for writing and experimenting with GPT-3.5. What I realized was that during this incubation period and the writing stage, I became less preoccupied with the mechanics of writing and more focused on accurately conveying my ideas. This heightened my motivation and enabled me to compose the text faster, compared to my previous experiences. Furthermore, my thinking and writing processes were both more efficient, streamlining the whole writing experience.

# Visual Comparisons of Reflections on using GPT-3.5 for Writing Articles / 1ˢᵗ Para:

Red: Removed | Green: Added

**Original to Results:**

1 | So, considering the effect of using GPT-3 on my motivation and practice of writing, I have mostly positive remarks. Because although the time that took to write this article might not be too short actually. I have written it in a matter of day, but it's not a very complicated article, it's not a very long article, so this might be quite trivial for some other people. But for me it was actuallThe use of GPT-3 to aid in the writing process had a largely positive effect on my motivation and productivity. Despite the fact that the article I wrote was not particularly complicated or lengthy, wthen I compared to my other writing practices, process of writing it was quite much more flowing. Because when I think about writing uid thand my ideas, I feel a bit overwhelmed when it's comes to that point that I need to write them down and turn them into ideas that are represented through text. I actually have been thinking for that project around maybe three weeks. I didn't somehow want to do it immediately because I wanted to think it through, see what kind of developments were happening. But throughout that process, I actually was really motivated to think about it. I didn't really think about writing that much. And this also continued for the day that I was writing this piece. I didn't really think too much about the hassle of writing, or thought too much about the words I would choousual practice. Before beginning to write, I had spent approximately three weeks mulling over the project and during the writing stage, I was less concerned with the mechanics of writing and more focused or anything like that, but more of the ideas that I want to represent. So it actually somehow helped with my thinking process and let me felt less overwhelmed. When it comes to the practice of writing, I also have realized I was more motivated and quite fast to put my ideas into these different chunks or somehow putting them into some kind of written form in a quite fast way. And even it might be still a good practice to do it and then start combining sentences and editing that, I usually feel quite much overwhelmed even to just put a chunk of text there thinking that I will struggle a lot to maken accurately representing my ideas. This increased my motivation to write, while also allowing me to compose text quickly. Moreover, this process also enabled me to think more quickly and efficiently, which further aided my writing. As a result, both my writ into a convincing, compelling and grammatically right and flowing text. Basically, both in my practice I was fed faster and more efficient, and also my thinking was more fast and efficient. So it really helped quite much in that sensg practice and thought processes were much more efficient, thereby streamlining my writing experience.

Difference (%): 98.54

**Results to Edit:**

> 1. The use of GPT-3 to aid in the writing process had a largely positive effect on my motivation and productivity. Despite the fact that the article I wrote was not particularly complicated or lengthy, the process of writing it was much more fluid than my usual practice. Before beginning to write, I had spent approximately three weeks mulling over the project. I did not act immediately so that the idea gets matures and I also wanted to follow the developments in the meantime. I also wanted to find two empty days where I can focus on writing and experimenting with GPT-3. Still, one thing I realized was that both during this incubation period and during the writing stage, I was less concerned with the mechanics of writing and more focused on accurately representing my ideas. This increased my motivation to write, while also allowing me to compose the text quicker than I could when I compared it to my previous writing experiences. Moreover, this process also enabled me to think more quickly and efficiently, which further aided my writing. As a result, both my writing practice and thought processes were much more efficient, thereby streamlining my writing experience.

**Difference (%): 30.68**

**Original Text 2 to Results 2:**

> 1. The use of GPT-3 to aid in the my writing process had a largely positive effect on my motivation and productivity. Despite the fact that Even though the article I wrote was not particularly complicated or lengthy, the process of writing it my writing process was much s more fluid other than my usual practice. Before beginn starting to write, I had spent approximately about three weeks mulling over the project. I did not act immediately so that, allowing the idea gets to matures and I also wanted to, follow the ing developments in the meantime. I also wanted to, and find ing two empty free days where I can focus on for writing and experimenting with GPT-3. Still, one thing What I realized was that both during this incubation period and during the writing stage, I was less concern became less preoccupied with the mechanics of writing and more focused on accurately represent convey ing my ideas. This increas heighten ed my motivation to write, while also allowing me to compose the text quicker than I could when I and enabled me to compose the text faster, compared it to my previous writing experiences. Moreover, this process also enabled me to think more quickly and efficiently, which further aided my writing. As a result, both my writing practice and thought Furthermore, my thinking and writing processes were muc both more efficient, thereby streamlining my the whole writing experience.

**Difference (%): 54.15**

**Results 2 to Final Edit:**

> 1. The use of GPT-3 to aid my writing process had a largely positive effect on my motivation and productivity. Even though the article I wrote was not particularly complicated or lengthy, my writing process was much smoother than usual. Before starting to write, I spent about three weeks mulling over the project, allowing the idea to mature, following developments, and finding two free days for writing and experimenting with GPT-3.5. What I realized was that during this incubation period and the writing stage, I became less preoccupied with the mechanics of writing and more focused on accurately conveying my ideas. This heightened my motivation and enabled me to compose the text faster, compared to my previous experiences. Furthermore, my thinking and writing processes were both more efficient, streamlining the whole writing experience.

**Difference (%): 0.24**

<div style="text-align:center">

**-END-**

</div>

# -NEW PARAGRAPH/SECTION-

**Section:** Reflections on using GPT-3.5 for Writing Articles
**Paragraph:** 2
**Method:** Talk Textualizer / Polisher

**Temperature Setting:** 0.9

**Prompt:**

> Turn the below text into academic writing, 2 paragraphs, clear, and rich academic language:

Original Text:

[transcribed from the spoken audio]

> Although the writing practice is of course an integral part of our jobs as academics. I believe across different disciplines there are different levels of training for that and there are different ways of tendency in terms of thinking about phenomena. Maybe some of the academics are more from social sciences fields such as philosophy or other fields of humanities might be more trained or more receptive to writing, while on the other hand people like me that are mostly from engineering areas or design might have less training and less abilities about writing. When I think about my total process of using GPT-3 for my, for textualizing my ideas, I can maybe compare it to my design practice. I'm a designer by training and I am quite adjusted to think about artifacts computers and machines to for example design and produce artifacts such as 3D CAD software, other types of visualization software and things such as 3D printers or other prototyping tools. And actually during my process I integrate all of those things into not just the imagination of what artifacts will be, but also process of how that will be produced. And in my opinion, using GPT-3 provided a similar freedom to me, which somehow put my thoughts off from the process of writing a little bit towards again what the ideas will look like or sound like. Of course, this still includes writing, there's still a lot of different practices as mentioned in the previous section for creating the text that's flowing, that has a kind of an understandable narrative, but at the same time relying on support from computers knowing that they will ease my job and will make my writing a lot more smoother. help me to feel relieved.

**Result 1:**

As academics, it is clear that writing is fundamental to our jobs. Different disciplines, however, have various levels of training and approaches regarding writing. For example, those in social sciences and humanities are often more extensively trained and more inclined towards writing, while academics such as those in engineering and design may have less proficiency and practice when it comes to written expression. This is why I was able to take advantage of and find a certain freedom in using GPT-3 for my writing process. As someone with a design background, I am accustomed to integrating various technologies, such as 3D CAD software, 3D printing, and prototyping, into tangible artifacts. Just as with these processes, GPT-3 allowed me to focus more on the ideas themselves rather than the writing. With the help of computers and GPT-3, the writing process became much more manageable and effortless. Nevertheless, there is still a need for multiple practices to form a coherent and understandable narrative.

Result 2:

Academic writing is a necessary skill for academics across all disciplines, however, the levels of training, experience, and receptivity to writing vary depends on the field of study. For example, those from social sciences and humanities may have more exposure to writing, while those from engineering and design may have less, such as myself. As I explored using GPT-3 to textualize my ideas, I found myself comparing it to my design practice. Having a background in design, I am accustomed to using tools such as CAD software, 3D printers and other prototyping tools to realize and tangible my ideas. In this regard, GPT-3 has provided a similar freedom as it enabled me to focus on the ideas I have, rather than the process of writing. Although basic writing skills are still required for the creation of a narrative, GPT-3 allows me to feel relieved of the burden of writing and create a smoother process.

**Edit:**

Mostly based on Result 2 and some parts from Result 1.

> Academic writing is a necessary skill for academics across all disciplines, however, the levels of training, experience, and receptivity to writing vary depends on the field of study. For example, those from social sciences and humanities may have more exposure to writing, while academics such as those in engineering and design may have less proficiency and practice when it comes to written expression. Especially in design field, writing compelling narratives are almost as important as other social sciences fields for expressing the value and the position of artifacts created or the design knowledge produces. I can definitely tell that I am more proficient in imagining artifacts and producing them compared to imagining stories and writing, and the manual nature of writing which rely on computational tools only for visual styling can be considered a bottleneck in my research process. However, as I explored using GPT-3 to textualize my ideas, I found myself comparing it to my design practice. Having a background in design, I am accustomed to using tools such as CAD software, 3D printers, electronic boards and other prototyping tools to realize tangible artifacts and these tools are also integral to my thinking process about those artifacts. In this regard, GPT-3 has provided a similar freedom since I relied on it similar to how I rely on computational design tools and it enabled me to focus on the ideas I have, rather than the daunting process of manual writing. Although basic writing skills are still required for the creation of a narrative, GPT-3 allows me to feel relieved of the burden of writing and create a smoother process.

**Prompt 2:**

> Copy-edit the following text, academic stlye, first person, clear, rich and flowing language:

**Original Text 2:**

Academic writing is a necessary skill for academics across all disciplines, however, the levels of training, experience, and receptivity to writing vary depends on the field of study. For example, those from social sciences and humanities may have more exposure to writing, while academics such as those in engineering and design may have less proficiency and practice when it comes to written expression. Especially in design field, writing compelling narratives are almost as important as other social sciences fields for expressing the value and the position of artifacts created or the design knowledge produces. I can definitely tell that I am more proficient in imagining artifacts and producing them compared to imagining stories and writing, and the manual nature of writing which rely on computational tools only for visual styling can be considered a bottleneck in my research process. However, as I explored using GPT-3 to textualize my ideas, I found myself comparing it to my design practice. Having a background in design, I am accustomed to using tools such as CAD software, 3D printers, electronic boards and other prototyping tools to realize tangible artifacts and these tools are also integral to my thinking process about those artifacts. In this regard, GPT-3 has provided a similar freedom since I relied on it similar to how I rely on computational design tools and it enabled me to focus on the ideas I have, rather than the daunting process of manual writing. Although basic writing skills are still required for the creation of a narrative, GPT-3 allows me to feel relieved of the burden of writing and create a smoother process.

**Result 3:**

Academic writing is an essential skill for scholars from all disciplines, though the level of training, experience, and receptivity to writing can vary depending on the field of study. For instance, those in the social sciences and humanities may have more practice with writing, while academics in engineering and design may have less proficiency and experience with written expression. In design, crafting compelling narratives is almost as important as in other social sciences in order to communicate the value and position of artifacts created or the design knowledge produced.

Personally, I find myself more adept at envisioning artifacts and bringing them to life than at imagining stories and writing, and the laborious nature of writing which largely relies on computational tools only for visual styling can be a bottleneck in my research process. However, upon exploring the use of GPT-3 to textualize my ideas, I was struck by the similarity to my design practice. With my background in design, I am accustomed to utilizing tools such as CAD software, 3D printers, electronic boards, and other prototyping tools for tangible artifacts, and these tools are integral to my thinking process surrounding them. Similarly, GPT-3 has provided me with a similar freedom as I have leaned on it in the same way I do with computational design tools; it has enabled me to focus on the ideas I have rather than the daunting task of manual writing. Although basic writing skills are still necessary for creating a narrative, GPT-3 has helped me to lighten the burden of writing and streamline the process.

**Final Edit:**

Academic writing is an essential skill for scholars from all disciplines, though the level of training, experience, and receptivity to writing can vary depending on the field of study. For instance, those in the social sciences and humanities may have more practice with and exposure writing, while academics in more applied fields such as engineering or design may have less proficiency and experience with written expression. Drawing on my own experience, in design research, crafting compelling narratives is almost as important as in other social sciences in order to communicate the value and position of artifacts created or the design knowledge produced. Personally, I find myself more adept at envisioning artifacts and bringing them to life (applied part of the design research) than at imagining stories and writing, and the laborious nature of writing process which incorporates computational tools only for visual styling can be a bottleneck in my research. However, upon exploring the use of GPT-3.5 to textualize my ideas, I recognized a similarity to my design practice. With my background in design, I am accustomed to utilizing tools such as CAD software, 3D printers, electronic boards, and other prototyping tools for creating tangible artifacts, and these tools are integral to my thinking process of designing things and makes it easier to conceptualize the process of thinking about artifacts and their production. Similarly, GPT-3.5 has provided me with a similar freedom as I have leaned on it in the same way I do with computational design tools; it has enabled me to focus on the ideas I have rather than the daunting task of manual writing. It also allowed me to work on my writing without distractions and interruptions. Although basic writing skills are still necessary for creating a narrative, GPT-3.5 has helped me to lighten the burden of writing and streamline the process.

## Visual Comparisons of Reflections on using GPT-3.5 for Writing Articles / 2nd Para:

Red: Removed | Green: Added

**Original to Results 2:**

1 | Although the writing practice is of course an integral part of our jobs as academics. I believe across different disciplines there are different levels of training for that and there are different ways of tendency in terms of thinking about phenomena. Maybe some of the academics are more from social sciences fields such as philosophy or other fields of cademic writing is a necessary skill for academics across all disciplines, however, the levels of training, experience, and receptivity to writing vary depends on the field of study. For example, those from social sciences and humanities might bay have more trained or more receptive to writing, while on the other hand people like me that are mostly exposure to writing, while those from engineering areas ornd design mightay have less training and less abilities about writing. When I think about my total process of, such as myself. As I explored using GPT-3 for my, forto textualizinge my ideas, I can maybefound myself compareing it to my design practice. I'm a designer by training and I am quite adjHaving a background in design, I am accustomed to think about artifacts computers and machines to for example design and produce artifactusing tools such as 3D CAD software, other types of visualization software and things such as 3D printers orand other prototyping tools. And actually during my process I integrate all of those things into not just the imagination of what artifacts will be, but also process of how that will be produced. And in my opinion, using to realize and tangible my ideas. In this regard, GPT-3 has provided a similar freedom to me, which somehow put my thoughts off from the process of writing a little bit towards again was it enabled me to focus on the ideas I have, rather than the ideas will look like or sound like. Of course, this still includes writing, there's stprocess of writing. Although basic writing skills a lot of different practices as mentioned in the previous section for creating the text that's flowing, that has a kind of an

understandable narrative, but at the same time relying on support from computers knowing that they will ease my job and will make my writing a lot more smoother. help me to feel relieved re still required for the creation of a narrative, GPT-3 allows me to feel relieved of the burden of writing and create a smoother process.

**Difference (%): 89.62**

**Results 2 to Edit:**

1. Academic writing is a necessary skill for academics across all disciplines, however, the levels of training, experience, and receptivity to writing vary depends on the field of study. For example, those from social sciences and humanities may have more exposure to writing, while academics such as those from in engineering and design may have less, such as myself. A proficiency and practice when it comes to written expression. Especially in design field, writing compelling narratives are almost as important as other social sciences fields for expressing the value and the position of artifacts created or the design knowledge produces. I can definitely tell that I am more proficient in imagining artifacts and producing them compared to imagining stories and writing, and the manual nature of writing which rely on computational tools only for visual styling can be considered a bottleneck in my research process. However, as I explored using GPT-3 to textualize my ideas, I found myself comparing it to my design practice. Having a background in design, I am accustomed to using tools such as CAD software, 3D printers, electronic boards and other prototyping tools to realize and tangible my ideas. In this regard, GPT-3 has provided a similar freedom as tangible artifacts and these tools are also integral to my thinking process about those artifacts. In this regard, GPT-3 has provided a similar freedom since I relied on it similar to how I rely on computational design tools and it enabled me to focus on the ideas I have, rather than the daunting process of manual writing. Although basic writing skills are still required for the creation of a narrative, GPT-3 allows me to feel relieved of the burden of writing and create a smoother process.

**Difference (%): 53.93**

**Original Text 2 to Results 3:**

1. Academic writing is a necessary skill for academics across n essential skill for scholars from all disciplines, however, though the levels of training, experience, and receptivity to writing can vary depending on the field of study. For example instance, those from in the social sciences and humanities may have more exposure to practice with writing, while academics such as those in engineering and design may have less proficiency and practice when it comes to written expression. Especially in design field, wri experience with written expression. In design, crafting compelling narratives are is almost as important as in other social sciences fields for expressing in order to communicate the value and the position of artifacts created or the design knowledge produces d. I can definitely tell that I am more proficient in imagi

2. Personally, I find myself more adept at envisioning artifacts and produc bringing them compared to to life than at imagining stories and writing, and the manual laborious nature of writing which rely largely relies on computational tools only for visual styling can be considered a bottleneck in my research process. However, as I upon explored using the use of GPT-3 to textualize my ideas, I found myself comparing it was struck by the similarity to my design practice. Having a With my background in design, I am accustomed to u stilizing tools such as CAD software, 3D printers, electronic boards, and other prototyping tools to realize for tangible artifacts, and these tools are also integral to my thinking process about those artifacts. In this regard surrounding them. Similarly, GPT-3 has provided me with a similar freedom since I relied on it similar to how I rely on as I have leaned on it in the same way I do with computational design tools and it; it has enabled me to focus on the ideas I have, rather than the daunting process task of manual writing. Although basic writing skills are still required necessary for the creation of ng a narrative, GPT-3 allows me to feel relieved of has helped me to lighten the burden of writing and c streate a smoom line ther process.

**Difference (%): 44.76**

**Results 3 to Final Edit:**

1. Academic writing is an essential skill for scholars from all disciplines, though the level of training, experience, and receptivity to writing can vary depending on the field of study. For instance, those in the social sciences and humanities may have more practice with and exposure writing, while academics in more applied fields such as engineering and or design may have less proficiency and experience with written expression. In design Drawing on my own experience, in design research, crafting compelling narratives is almost as important as in other social sciences in order to communicate the value and position of artifacts created or the design knowledge produced.

2. Personally, I find myself more adept at envisioning artifacts and bringing them to life (applied part of the design research) than at imagining stories and writing, and the laborious nature of writing which largely relies on process which incorporates computational tools only for visual styling can be a bottleneck in my research process. However, upon exploring the use of GPT-3.5 to textualize my ideas, I was struck by the recognized a similarity to my design practice. With my background in design, I am accustomed to utilizing tools such as CAD software, 3D printers, electronic boards, and other prototyping tools for creating tangible artifacts, and these tools are integral to my thinking process surrounding them of designing things and makes it easier to conceptualize the process of thinking about artifacts and their production. Similarly, GPT-3.5 has provided me with a similar freedom as I have leaned on it in the same way I do with computational design tools; it has enabled me to focus on the ideas I have rather than the daunting task of manual writing. It also allowed me to work on my writing without distractions and interruptions. Although basic writing skills are still necessary for creating a narrative, GPT-3.5 has helped me to lighten the burden of writing and streamline the process.

**Difference (%): 23.07**

# - END -

# -NEW PARAGRAPH/SECTION-

**Section:** Reflections on using GPT-3.5 for Writing Articles
**Paragraph:** 3
**Method:** Talk Textualizer

**Temperature Setting:** 0.7

**Prompt:**

> Turn the below text into academic writing, clear, first person, and rich academic language:

**Original Text:**

[transcribed from the spoken audio]

> However, something I thought was also be that I would be more effective in terms of to spend time on writing. This hasn't been true to some extent because Still, the work I have put in, for example, writing a chunk of text or talking about them and then spending time with retrials until I reach something that satisfies me actually took quite a long time. For example, most probably I have put the main material of introduction I something around 10 minutes while getting a result that would satisfy me took around one and a half hours of retries with GPT-3. In that sense, although I think in total it took less time than I would spend for creating the whole text because of less distractions I had while using GPT-3, I believe still, I believe that still, and authors need to spend quite a lot of time on writing, on polishing, on trying, and when we also include the current slow nature of GPT-3 due to its servers being too overburdened, I wouldn't see it as a fast solution for writing more than a tool that makes writing more effective and actually incorporates new ways of writing which might be more suitable to people with different levels of skills.

**Result:**

> I have found that I have not been as effective with my writing as I had anticipated. Despite the time I have invested, such as writing a section of text and then making multiple revisions until I reach a satisfactory outcome, this process has taken quite some time. For instance, it took me approximately 10 minutes to write the primary material of an introduction while the subsequent revisions took around 90 minutes, utilizing GPT-3. Although I believe I was able to write more quickly due to the lack of distractions while using GPT-3, I still believe that authors need to dedicate a substantial amount of time to writing and refining their work. In addition, the current slow speed of GPT-3 due to the overwhelming demand on its servers means it cannot be considered a quick solution for writing, but rather a tool that makes writing more effective and introduces new writing techniques suitable for different skill levels.

**Final Edit:**

```
I have also found that I have not been as time-efficient with my writing as
I had anticipated. Writing a section of text and then making multiple
revisions until I reach a satisfactory outcome has taken quite some time.
For instance, it took me approximately 10 minutes to write the primary
material of an introduction while the subsequent revisions took around 90
minutes which is quite long for such a short section. Although I believe I
was able to write more quickly due to the lack of distractions which happens
in my usual writing practice because of the interruptions in my thought
process, I still believe that authors need to dedicate a substantial amount
of time in the process. For creating the report which shows the comparisons
between the original text, outcomes of GPT-3.5 and final edited text, I
spent around 5 more hours. In addition, the current slow speed of GPT-3.5
due to the overwhelming demand on its servers means it cannot be considered
a quick solution for writing, but rather a tool that makes writing process
smoother and introduces new writing techniques suitable for different skill
levels.
```

## Visual Comparisons of Reflections on using GPT-3.5 for Writing Articles / 3rd Para:

Red: Removed | Green: Added

**Original to Result:**

1  However, something I thought was also be that I would be more effective in terms of to spend time on writing. This hasn't been true to some extent because I have found that I have not been as effective with my writing as I had anticipated. Despite the Still, the work I have put in, for example, me I have invested, such as writing a chunk section of text or talking about them and then spending time with retrials until I reach something that satisfies me actually took and then making multiple revisions until I reach a satisfactory outcome, this process has taken quite a long some time. For example, most probably I have put the main instance, it took me approximately 10 minutes to write the primary material of an introduction I something around 10 minutes while getting a result that would satisfy me took around one and a half hours of retries with GPT-3. In that sense, although I think in total it took less time than I would spend for creating the whole text because of less while the subsequent revisions took around 90 minutes, utilizing GPT-3. Although I believe I was able to write more quickly due to the lack of distractions I had while using GPT-3, I believe still, I believe that still, and authors need to spend quite a lo dedicate a substantial amount of time t on writing, on polishing, on trying, and when we also include and refining their work. In addition, the current slow nature speed of GPT-3 due to its servers being too overburdened, I wouldn't see it as a fast the overwhelming demand on its servers means it cannot be considered a quick solution for writing more than, but rather a tool that makes writing more effective and actually incorporates new ways of writing which might be more suitable to people with different levels of skil introduces new writing techniques suitable for different skill level s.

**Difference (%): 85.65**

**Result to Final Edit:**

1. I have ~~also~~ found that I have not been as ~~effective~~time-efficient with my writing as I had anticipated. ~~Despite the time I have invested, such as w~~Writing a section of text and then making multiple revisions until I reach a satisfactory outcome~~, this process~~ has taken quite some time. For instance, it took me approximately 10 minutes to write the primary material of an introduction while the subsequent revisions took around 90 minutes~~, utilizing GPT-3~~ which is quite long for such a short section. Although I believe I was able to write more quickly due to the lack of distractions whi~~le using GPT-3, I still believe that authors need to dedicate a substantial amount of time to writing~~ch happens in my usual writing practice because of the interruptions in my thought process, I still believe that authors need to dedicate a substantial amount of time in the process. For creating the report which shows the comparisons between the original text, outcomes of GPT-3.5 and ~~refin~~ing their work~~al edited text~~, I spent around 5 more hours. In addition, the current slow speed of GPT-3.5 due to the overwhelming demand on its servers means it cannot be considered a quick solution for writing, but rather a tool that makes writing ~~more effective~~process smoother and introduces new writing techniques suitable for different skill levels.

**Difference (%): 46.52**

# -END-

# -NEW PARAGRAPH/SECTION-

**Section:** Reflections on using GPT-3.5 for Writing Articles
**Paragraph:** 4
**Method:** Talk Textualizer

**Temperature Setting:** 0.7

**Prompt:**

> Copy-edit the following text, academic style, first person, clear, rich and flowing language:

Original Text:

[transcribed from the spoken audio]

> Regarding the quality of this text, I believe that the quality was variable across different sections. I could reflect on this especially during making the last round of revisions to text from beginning to end after all written content was placed. In the process, first editing was done immediately after the results were gotten from GPT-3 and the last round of revisions were done after the whole piece is written. Some parts of the manuscript were read more plain and unengaging, while some other parts were easier to read and more compelling. I would most probably like to do a couple of more rounds of revisions to have a better consistency of the tone and a more flowing narrative, however, I intentionally left the manuscript in the current state so that it would serve as a demonstration of what can be produced with GPT-3 in a relatively short time and with minimal editing. However, I believe that several more rounds of revision would be necessary for creating an academic article which is consistent in language style and with a nice narrative flow.

**Result:**

> I believe that the quality of this text is variable across different sections. As I made the last round of revisions, I noticed this especially when reading through the entire piece. Initially, I edited the text immediately after receiving the results from GPT-3, and then I made further revisions after the entire piece was written. Some sections were plain and unengaging, while others were easier to read and more compelling. I could have done a few more rounds of revisions to ensure a consistent tone and a more flowing narrative, but I left the manuscript as it is to demonstrate what can be produced with GPT-3 in a short time and with minimal editing. However, I believe that several more rounds of revision would be necessary to create an academic article with a consistent language style and a nice narrative flow.

Final Edit:

> I believe that the quality of this text is variable across different sections. As I made the last round of revisions, I noticed this especially when reading through the entire piece. Initially, I edited the text immediately after receiving the results from GPT-3.5, and then I made further revisions after the entire piece was written. Some sections were plain and unengaging, while others were easier to read and more compelling. I could have done a few more rounds of revisions to ensure a consistent tone and a more flowing narrative, but I left the manuscript as it is to demonstrate what can be produced with GPT-3.5 in a short time (around a day) and with minimal editing. However, I believe that several more rounds of revision would be necessary to create an academic article with a consistent language style and a solid narrative flow.

**Visual Comparisons of Reflections on using GPT-3.5 for Writing Articles / 4th Para:**

Red: Removed | Green: Added

**Original to Result:**

1. ~~Regarding~~ I believe that the quality of this text~~, I believe that the quality wa~~ is variable across different sections. ~~I could reflect on this especially during making the last round of revisions to text from beginning to end after~~ As I made the last round of revisions, I noticed this especi~~ally w~~ritten content was pla~~h~~en reading through the entire pie~~ced~~. In the process, first editing was done~~itially, I edited the text immediately after ~~receiving~~ the results ~~were gotten~~ from GPT-3, and the~~ last round of~~n I made further revisions ~~were done~~ after the who~~entir~~e piece ~~i~~wa~~s~~ written. Some ~~parts of the manuscript were read mo~~sections we~~r~~e plain and unengaging, while ~~some other part~~others were easier to read and more compelling. I w~~ould most probably like to do a couple of~~ more rounds of revisions to ~~have a better consistency of the tone and a more flowing narrative, however, I intentionally left the manuscript in the current state so that it would serve as a~~have done a few more rounds of revisions to ensure a consistent tone and a more flowing narrative, but I left the manuscript as it is to demonstrat~~ion of~~e what can be produced with GPT-3 in a ~~relatively~~ short time and with minimal editing. However, I believe that several more rounds of revision would be necessary ~~for~~to creat~~ing~~e an academic article w~~hich is~~ith a consistent in language style and ~~with~~ a nice narrative flow.

**Difference (%):** 64.78

**Result to Final Edit:**

1. I believe that the quality of this text is variable across different sections. As I made the last round of revisions, I noticed this especially when reading through the entire piece. Initially, I edited the text immediately after receiving the results from GPT-3.5, and then I made further revisions after the entire piece was written. Some sections were plain and unengaging, while others were easier to read and more compelling. I could have done a few more rounds of revisions to ensure a consistent tone and a more flowing narrative, but I left the manuscript as it is to demonstrate what can be produced with GPT-3.5 in a short time (around a day) and with minimal editing. However, I believe that several more rounds of revision would be necessary to create an academic article with a consistent language style and a ~~nice~~ solid narrative flow.

Difference (%): 3.30

# -END-

# -NEW PARAGRAPH/SECTION-

**Section:** Ways of Using Methods
**Paragraph:** Whole Section
**Method:** Talk Textualizer

**Temperature Setting:** 0.2

**Prompt:**

> Turn the below text into academic writing, clear, first person, and rich academic language:

Original Text:

[transcribed from the spoken audio]

> When it comes to methods I have used while writing the paper, I think I rather used mostly the Chunk Stylist and Taulk Textualizer. The reason behind that might be that they actually allowed me to at least formalize my ideas in a more concrete way so that I didn't need to rely on some ideas that are not mine, that was generated by GPT-3 For example, when I used bullet to paragraph or research body, I needed to make maybe a few more trials and most of needed to edit them more substantially in others because it created some sentences which are not actually based on facts and also when I wasn't quite clear enough what I want to convey in actual fields of text with the ideas that I wasn't thinking or even kind of really contrasting with the things that I was thinking. So therefore, in my opinion, to write a formal article among the methods that I have introduced above. The chunk stylist and talk textualizer and to some extent bullet to paragraph might be better than using for example ResearchBuddy which creates a lot of more spontaneous ideas. for example for defining what GPT-3 is and give it a source to base it to. The result that I got was not really based on that source when checked the source. I tried to understand if the summary that GPT-3 made really aligned with what was written in the source and the source was kind of technical so I might have not understood it fully, but since I could not find enough information to support what GPT-3 produced, I needed to remove most of the things and just kind of left the base comments that I know were true. when it comes to polisher, I have mostly used it for polishing the prompts that I edited but of course it can be also used by more proficient writers after for copy editing their text.

**Result:**

I utilized the Chunk Stylist and Taulk Textualizer predominantly when writing my paper. This decision was formed by their ability to formalize my ideas in a more concrete manner, rather than relying on material generated by GPT-3. For example, when using Bullet to Paragraph or Research Body, I had to make several attempts and perform substantial editing to avoid sentences not based on facts. Thus, I believe the aforementioned methods, particularly the Chunk Stylist and Taulk Textualizer, are more suitable for writing a formal article than ResearchBuddy, which produces more spontaneous ideas. When I tried to gain a better understanding of GPT-3, I found that the summary generated was not reflective of the source material. As I was unable to identify enough supporting information to back up GPT-3's output, I had to remove most of the material and leave only the core points that I knew were accurate. Polisher was mainly used to refine the prompts that I had edited; however, it can be employed by more proficient writers for copy editing their text.

**Final Edit:**

I utilized the methods of Chunk Stylist and Talk Textualizer predominantly when writing this article. I was inclined to use them because of their ability to formalize my own ideas, rather than relying on spontaneously generated material by GPT-3.5. When using Bullet to Paragraph or Research Buddy, I had to make several attempts and perform substantial editing to avoid sentences not based on facts. Thus, I believe the Chunk Stylist and Talk Textualizer are more suitable for writing a formal article than methods relying on big portion of generated text such as Research Buddy. For example, when I prompted GPT-3.5 to give me the description of GPT-3 based on (Brown et al., 2020), I found that the summary generated was not reflective of the source material (the document was mostly technical and required good understanding of NLP, so I could not assess the accuracy of information). As I was unable to identify enough supporting information to back up GPT-3.5's output, I had to remove most of the material and leave only the core points that I knew were accurate. I used Polisher mainly to refine the results that I had edited; however, it can be employed by proficient writers for copy-editing their text.

## Visual Comparisons of Ways of Using Methods / Whole Section:

Red: Removed | Green: Added

**Original to Result:**

1. When it comes to methods I have used while writing the paper, I think I rather used mostly the Chunk Stylist and Taulk Textualizer. The reason behind that might be that they actually allowed me to at least I utilized the Chunk Stylist and Taulk Textualizer predominantly when writing my paper. This decision was formed by their ability to formalize my ideas in a more concrete way so that I didn't need to manner, rather than relying on some ideas that are not mine, that was material generated by GPT-3. For example, when I used busing Bullet to pParagraph or rResearch bBody, I needehad to make maybe a few more triseverals and most of needed to edit them mttempts and perforem substantially in others because it created some sentences which are not actually based on facts and also when I wasn't quite clear enough what I want to convey in actual fields of text with the ideas that I wasn't thinking or even kind of really contrasting with the things that I was thinking. So therefore, in my opinion, to write a formal article among the methods that I have introduced above. T editing to avoid sentences not based on facts. Thus, I believe the aforementioned methods, particularly the cChunk sStylist and taTaulk tTextualizer, and to some extent bullet to paragraph might be better than usre more suitable for writing a for examplemal article than ResearchBuddy, which creates a lot of produces more spontaneous ideas. for example for defining what GPT-3 is and give it a source to base it to. The result that I got was not really based on that source when checked the source. I tried to understand if the summary that GPT-3 made really aligned with what was written in the source and the source was kind of technical so I might have not understood it fully, but since I could not find enough information to support what GPT-3 produced, I needed to remove most of the things and just kind of left the base commeWhen I tried to gain a better understanding of GPT-3, I found that the summary generated was not reflective of the source material. As I was unable to identify enough supporting information to back up GPT-3's output, I had to remove most of the material and leave only the core points that I knew were true. when it comes to polisher, I have mostly used it for polishaccurate. Polisher was mainly used to refinge the prompts that I had edited but of course; however, it can be also usemployed by more proficient writers after for copy editing their text.

**Difference (%): 85.64**

### Result to Final Edit:

1. I utilized the methods of Chunk Stylist and Taulk Textualizer predominantly when writing my paper. This decision was formed bythis article. I was inclined to use them because of their ability to formalize my own ideas in a more concrete manner, rather than relying on material generated by GPT-3. For example, w, rather than relying on spontaneously generated material by GPT-3.5. When using Bullet to Paragraph or Research Bouddy, I had to make several attempts and perform substantial editing to avoid sentences not based on facts. Thus, I believe the aforementioned methods, particularly the Chunk Stylist and Taulk Textualizer, are more suitable for writing a formal article than ResearchBuddy, which produces more spontaneous ideas. When I tried to gain a better understanding of GPT-3, I found that the summary generated was not reflective of the source materialmethods relying on big portion of generated text such as Research Buddy. For example, when I prompted GPT-3.5 to give me the description of GPT-3 based on (Brown et al., 2020), I found that the summary generated was not reflective of the source material (the document was mostly technical and required good understanding of NLP, so I could not assess the accuracy of information). As I was unable to identify enough supporting information to back up GPT-3.5's output, I had to remove most of the material and leave only the core points that I knew were accurate. I used Polisher was mainly used to refine the prompresults that I had edited; however, it can be employed by more proficient writers for copy-editing their text.

**Difference (%): 38.46**

# - END -

# -NEW PARAGRAPH/SECTION-

**Section:** Transparency, Agency and Biases
**Paragraph:** Whole Section
**Method:** Talk Textualizer

**Temperature Setting:** 0.7

**Prompt:**

> Turn the below text into academic writing, clear, first person, and rich academic language:

Original Text:

[transcribed from the spoken audio]

> As we have discussed in the background, one of the biggest concerns of incorporating GPT-3 into the writing process is the concerns about plagiarism and transparency. And that's also, according to my trial, a very complicated issue. On the one hand, some of the prompts and results that I got is not really different from using, for example, tools like Grammarly or even the spell or grammar corrector of Microsoft Word. In some other cases, I would tell that I wasn't sure what my agency of the written text. Of course, maybe this experience was more similar to using a professional copy editing service, but maybe seeing it that fast and also not knowing where this words and sentences were scraped from actually created mixed feelings about the content I was about to use. Of course it is not mine to decide if it is ethical to use these tools in academic writing. In my opinion, authors should follow some transparency measures. In this article I actually documented all the prompts I have used and results and also the final edited versions of those. And in some cases I think it is very needed. For example By using the methods of talk textualizer or let's say the research body or the bullet paragraph, it is very easy to see that the content and the language and the tone of the writing is heavily influenced and modified by GPT-3. In these kind of situations, I think it is of critical importance to be transparent and show and clearly demonstrate what was the process of this writing and how and what it is transformed. On the other hand, for example, Nature, when in their editorial mentioned that authors should mention that they have used in GPT-3 in their writing, but in my opinion that might not be enough. If GPT-3 have only been used for grammatical correcting, spelling and somehow minor copyediting reasons, maybe a note in the acknowledgement might be enough. But if some other methods that I described above have been used, especially a talk textualizer, or bullet 2 paragraph, or... or... or a research body, I think a more substantial reporting, maybe in forms of an external link or an appendix. is needed to ensure transparency. And of course it would create extra burden for reviewers to understand what was the extent of chains. So we might need better tools that would demonstrate and highlight for example changes where it was heavy or even somehow trying to provide transparency by trying to show the source of source that has been used to generate this text.

Result:

In our discussion of the background, one of the primary concerns of incorporating GPT-3 into the writing process is the potential for plagiarism and a lack of transparency. My own trial of GPT-3 revealed a complex issue. In some instances, the prompts and results I obtained were not dissimilar from those produced by tools like Grammarly or the spell and grammar corrector of Microsoft Word. In other cases, however, I was unsure of the agency of the written text. This experience was arguably similar to using a professional copy editing service, yet the speed of the process and the lack of knowledge about where the words and sentences derived from caused me to feel uneasy about using the content.

It is not for me to decide whether it is ethical to use these tools in academic writing. Nevertheless, I believe that authors should still practice transparency. In this article, I have documented all the prompts, results, and final edited versions I used. In some cases, this is essential; for example, when using methods such as Talk Textualizer, Bullet Paragraph, or Research Body, the heavy influence of GPT-3 on the content, language, and tone of the writing is readily apparent. In such situations, it is of critical importance to be transparent and clearly demonstrate the process of the writing and how it has been transformed.

On the other hand, when Nature's editorial suggested that authors mention their use of GPT-3 in their writing, I believe this may not be enough. If GPT-3 has only been used for minor copyediting purposes, then a note in the acknowledgement may suffice. However, if other methods have been used, such as Talk Textualizer or Bullet Paragraph, a more substantial reporting—perhaps in the form of an external link or an appendix—is necessary to ensure transparency. Additionally, better tools may be required to demonstrate the extent of changes and highlight where heavy modifications have been made, as well as to provide transparency by showing the source from which the text was generated.

**Final Edit:**

One of the primary concerns of incorporating GPT-3.5 into the writing process is the potential for plagiarism and a lack of transparency ("Next Chapter in Artificial Writing," 2020). In my own trials, some instances, the results I obtained were not dissimilar from those produced by tools like Grammarly or the spell and grammar checker of Microsoft Word. In other cases, however, I was unsure of my agency over the written text. This experience was arguably similar to using a professional copy-editing service where I need to carefully check the text and make sure that the intended meanings are retained. However, the speed of the process and the lack of knowledge about where the words and sentences derived from caused me to feel uneasy about using the content. Overall, looking at the full text, the difference between the original texts I prompted to GPT-3.5 and the outcomes I got was 70.45%, while the difference between outcomes and the final edited version was 28%. The biggest difference between an original text chunk and an outcome was 98.54% (almost the whole text were changed), the biggest difference between the outcome and the final edited part was 71.44% and the smallest was 3.3% (copy-pasted to this manuscript almost without a change). You can see the detailed report in the supplemented document.

This variability in author's agency over the result produced GPT-3.5 require to develop practices for transparency for articles where GPT-3.5 (or other LLM) incorporated. In this article, I have documented all the prompts, results, and final edited versions I used. In some cases, this is essential; for example, when using methods such as Talk Textualizer, Bullet to Paragraph, or Research Buddy, the heavy influence of GPT-3.5 on the content, language, and tone of the writing is readily apparent. In such situations, it is of critical importance to be transparent and clearly demonstrate the process of the writing and how it has been transformed.

In their recent editorial, Nature suggested that authors must mention their use of GPT-3.5 in their writing ("Tools Such as ChatGPT Threaten Transparent Science; Here Are Our Ground Rules for Their Use," 2023), but I believe this may not be enough. If GPT-3.5 has only been used for minor copyediting purposes, then a note in the acknowledgement may suffice. However, if other methods have been used, such as Talk Textualizer or Bullet to Paragraph, a more substantial reporting—perhaps in the form of an external link or an appendix—is necessary to ensure transparency, which also has been used in other contexts such as classroom assignments (Fyfe, 2022). This would also ensure that the authors would thoroughly check the content making sure that that their content do not consolidate racial biases or any other radicalized political ideas unintentionally. Additionally, better tools may be required to demonstrate the extent of changes and highlight where heavy modifications have been made, as well as to provide transparency by showing the source from which the text was generated.

# Visual Comparisons of Ways of Using Methods / Whole Section:

Red: Removed | Green: Added

**Original to Result:**

1. As we have In our discussed in ion of the background, one of the biggest primary concerns of incorporating GPT-3 into the writing process is the concerns about potential for plagiarism and a lack of transparency. And that's also, according to my trial, a very complicated issue. On the one hand, some of the prompts and results that I got is not really different from using, for example, My own trial of GPT-3 revealed a complex issue. In some instances, the prompts and results I obtained were not dissimilar from those produced by tools like Grammarly or even the spell or and grammar corrector of Microsoft Word. In some other cases, I would tell that however, I wasn't un sure what my of the agency of the written text. Of course, maybe t This experience was more arguably similar to using a professional copy editing service, but maybe yet the spee ing it that fast and also not knowing d of the process and the lack of knowledge about where th ise words and sentences were scraped from actually created mixed feelings derived from caused me to feel uneasy about using the content I was about to use. Of course it is not min.

2. It is not for me to decide if whether it is ethical to use these tools in academic writing. In my opinion, Nevertheless, I believe that authors should follow som still practice transparency measures. In this article, I actually have documented all the prompts I have used and, results, and also the final edited versions of tho I used. And i In some cases I thi nk it is very needed. F s is essential; f or example By, when using the methods of t such as Talk t Textualizer or let's say the r, Bullet Paragraph, or Research b Body or, the bullet paragraph, it is very easy to see that heavy influence of GPT-3 on the content and the, language, and the tone of the writing is h rea vd ily influenced and modified by GPT-3 apparent. In these kind of such situations, I think it is of critical importance to be transparent and show and clearly demonstrate what was the process of th ise writing and how and what it is it has been transformed.

3. On the other hand, for example, Nature, when in their when Nature's editorial mention suggest ed that authors should mention th at they have used in eir use of GPT-3 in their writing, but in my opinion that might I believe this may not be enough. If GPT-3 ha ves only been used for grammatical correcting, spelling and somehow minor copyediting reasons, maybe purposes, then a note in the acknowledgement m ight be enough. But if some other methods that I described above ay suffice. However, if other methods have been used, especially a t such as Talk t Textualizer, or b Bullet 2 p Paragraph, or... or... or a research body, I think a more substantial reporting, maybe in —perhaps in the form s of an external link or an appendix. —is nee eded cessary to ensure transparency. And of course it would create extra burden for reviewers to understand what was the extent of chains. So we might need better tools that would demonstrate and highlight for example changes where it was heavy or even somehow trying dditionally, better tools may be required to demonstrate the extent of changes and highlight where heavy modifications have been made, as well as to provide transparency by trying to show ing the source of source tha from which the text h was been used to generate this text. generated.

Difference (%): 63.58

**Result to Final Edit:**

1. One of the primary concerns of incorporating GPT-3.5 into the writing process is the potential for plagiarism and a lack of transparency. My own trial of GPT-3 revealed a complex issue. In my own trials, some instances, the results I obtained were not dissimilar from those produced by tools like Grammarly or the spell and grammar checker of Microsoft Word. In other cases, however, I was unsure of my agency over the written text. This experience was arguably similar to using a professional copy-editing service, where I need to carefully check the text and make sure that the intended meanings are retained. However, the speed of the process and the lack of knowledge about where the words and sentences derived from caused me to feel uneasy about using the content.

2. Overall, looking at the full text, the difference between the original texts I prompted to GPT-3.5 and the outcomes I got was 70.45%, while the difference between outcomes and the final edited version was 28%. The biggest difference between an original text chunk and an outcome was 98.54% (almost the whole text were changed), the biggest difference between the outcome and the final edited part was 71.44% and the smallest was 3.3% (copy-pasted to this manuscript almost without a change). You can see the detailed report in the supplemented document.

3. This variability in author's agency over the result produced GPT-3.5 require to develop practices for transparency for articles where GPT-3.5 (or other LLM) incorporated. In this article, I have documented all the prompts, results, and final edited versions I used. In some cases, this is essential; for example, when using methods such as Talk Textualizer, Bullet to Paragraph, or Research Buddy, the heavy influence of GPT-3.5 on the content, language, and tone of the writing is readily apparent. In such situations, it is of critical importance to be transparent and clearly demonstrate the process of the writing and how it has been transformed.

4. In their recent editorial, Nature suggested that authors must mention their use of GPT-3.5 in their writing, ("Tools Such as ChatGPT Threaten Transparent Science; Here Are Our Ground Rules for Their Use," 2023), but I believe this may not be enough. If GPT-3.5 has only been used for minor copyediting purposes, then a note in the acknowledgement may suffice. However, if other methods have been used, such as Talk Textualizer or Bullet to Paragraph, a more substantial reporting—perhaps in the form of an external link or an appendix—is necessary to ensure transparency, which also has been used in other contexts such as classroom assignments (Fyfe, 2022). This would also ensure that the authors would thoroughly check the content making sure that that their content do not consolidate racial biases or any other radicalized political ideas unintentionally. Additionally, better tools may be required to demonstrate the extent of changes and highlight where heavy modifications have been made, as well as to provide transparency by showing the source from which the text was generated.

`Difference (%): 47.33`

**- END -**

# -NEW PARAGRAPH/SECTION-

**Section:** Ecological Sustainability
**Paragraph:** Whole Section
**Method:** Talk Textualizer

**Temperature Setting:** 0.7

**Prompt:**

> Turn the below text into academic writing, clear, first person, and rich academic language:

Original Text:

[transcribed from the spoken audio]

> Another concern of using GPT-3 is around ecological sustainability. It's also a very valid concern. I have created around 500 requests in the scope of writing this article. And one behavior I have also observed on myself was that I was carelessly reprompting it for the same paragraph until I reach a satisfactory result. This article costed around $3 according to the calculation of OpenAI's usage and I am not aware of the carbon footprint I have produced in the scope of writing this article. On one hand, when we think GPT-3 and similar language models as in integrated parts of Academy It might have a substantial negative effect on the energy consumption Because on the other alternative The toad process would be... the thought process of writing would be facilitated by my brain and body which might have... spent less resources. On the other hand, I'm pretty sure that writing this piece would take much longer than I did. But as I have described, not because GP3 is very time efficient, but because I am more motivated and less prone to distractions while using that. I might have also then spend less resources such as electricity of the computer or the office spaces, let's say, or other facilities of my university. Another advantage is usually my writing journeys end in the middle of night after hours of procrastination because I'm not motivated or my brain works in a way that postpone writing tasks that might have, for example, a benefit on my mental health on the long term. And if that would apply to a wide range of researchers, create a burden of trying to write pieces in relatively limited and short time, which might be a positive outcome. Thus, although it is very apparent that, according to the previous studies, using artificial intelligence and language models prevalent in academia would definitely have a negative impact On ecology, each of us should care and then regulate our behavior accordingly. For example, make sure that we use it with least prompts as possible. I would also invite to think positive impact of using those tools beyond just being more efficient in writing tasks.

**Result**

> The use of GPT-3 raises ecological sustainability concerns. In writing this article, I created approximately 500 requests and often carelessly reprompted it for the same paragraph until reaching a satisfactory result. According to OpenAI's calculations, this article costed around $3, but I am unaware of the carbon footprint created in doing so. On the one hand, the integration of GPT-3 and similar language models into academia could have a substantial negative effect on energy consumption in comparison to the thought process facilitated by my brain and body, which could have resulted in less resource expenditure. On the other hand, my writing process could have been more efficient with the use of GPT-3; however, I am more motivated and less prone to distraction while using it, potentially reducing electricity costs of my computer, office space, and other university facilities. Additionally, my writing journeys often end in the middle of the night, which is arguably beneficial for my mental health in the long-term. Thus, although using artificial intelligence and language models in academia could have a negative impact on ecology, regulating our behavior accordingly, such as using it with least prompts as possible, may offset these effects. Additionally, we should consider the positive impacts of using these tools beyond just being more efficient in writing tasks.

**Final Edit:**

> The use of GPT-3.5 and similar tools raises ecological sustainability concerns (Iris, 2023). In writing this article, I created approximately 500 requests and often carelessly reprompted GPT-3.5 for the same paragraph until reaching a satisfactory result. According to OpenAI's calculations, this article costed around $2.5, but I am unaware of the carbon footprint created in doing so.
>
> On the one hand, the integration and normalization of GPT-3.5 and similar language learning models into academia could have a substantial negative effect on energy consumption in comparison to the thought process facilitated by the brain and body, which could have resulted in less resource expenditure. On the other hand, my writing process has been more efficient with the use of GPT-3.5 – not because it is time-efficient as a tool but makes me more motivated and less prone to distraction while using it – potentially reducing electricity costs spent by my computer or office space. Additionally, my writing journeys often extend to the middle of the night, which is arguably not beneficial for my physical and mental health in the long-term, lowering the consequences of mental burden caused by the pressure and stress of writing preventing overworking, a known problem in academia (Urbina-Garcia, 2020).
>
> Using tools like GPT-3.5 in academia could have a significant negative impact on ecology and as academics we are responsible of regulating our behavior accordingly, such as using it with least prompts as possible. Still, we should consider the positive impacts of using these tools beyond just being more efficient in writing tasks, such as its impact on more efficient utilization of resources and improving the wellbeing of academics.

**Visual Comparisons of Ecological Sustainability / Whole Section:**

Red: Removed | Green: Added

Original to Result:

1  The use of GPT-3 raises ecological sustainability concerns. When writing this article, I created approximately 500 requests and often carelessly reprompted it for the same paragraph until reaching a satisfactory result. According to OpenAI's calculation of OpenAI's usage, this article costed around $3, but I am not aware of the carbon footprint created in doing so. On the one hand, integration of GPT-3 and similar language models into academia could have a substantial negative effect on energy consumption in comparison to the thought process of writing facilitated by my brain and body, which could have resulted in less resource expenditure. On the other hand, my writing process could have been more efficient with the use of GPT-3; however, I am more motivated and less prone to distraction while using that. I might have also then spend less resources such as it, potentially reducing electricity costs of my computer or other university facilities. Additionally, my writing journeys often end in the night, which is arguably beneficial for my mental health in the long term. Thus, although using artificial intelligence and language models prevalent in academia could definitely have a negative impact on ecology, each of us should care and then regulate our behavior accordingly, such as using it with least prompts as possible, may offset these effects. Additionally, we should consider the positive impact of using those tools beyond just being more efficient in writing tasks.

Difference (%): 72.69

**Result to Final Edit:**

1. The use of GPT-3.5 and similar tools raises ecological sustainability concerns (Iris, 2023). In writing this article, I created approximately 500 requests and often carelessly reprompted GPT-3.5 for the same paragraph until reaching a satisfactory result. According to OpenAI's calculations, this article costed around $2.5, but I am unaware of the carbon footprint created in doing so.

2. On the one hand, the integration and normalization of GPT-3.5 and similar language learning models into academia could have a substantial negative effect on energy consumption in comparison to the thought process facilitated by the brain and body, which could have resulted in less resource expenditure. On the other hand, my writing process has been more efficient with the use of GPT-3.5 - not because it is time-efficient as a tool but makes me more motivated and less prone to distraction while using it, potentially reducing electricity costs spent by my computer, office space, and other university facilities. Additionally, my writing journeys often extend into the middle of the night, which is arguably not beneficial for my physical and mental health in the long-term. Thus, although using artificial intelligence and language models, lowering the consequences of mental burden caused by the pressure and stress of writing preventing overworking, a known problem in academia (Urbina-Garcia, 2020).

3. Using tools like GPT-3.5 in academia could have a significant negative impact on ecology, and as academics we are responsible of regulating our behavior accordingly, such as using it with least prompts as possible. Still, we should consider the positive impacts of using these tools beyond just being more efficient in writing tasks, such as its impact on more efficient utilization of resources and improving the wellbeing of academics.

**Difference (%): 36.65**

# - END -

# -NEW PARAGRAPH/SECTION-

**Section:** Additional tools that can help with GPT based academic writing
**Paragraph:** Whole Section
**Method:** Talk Textualizer

**Temperature Setting:** 0.7

**Prompt:**
> Turn the below text into academic writing, clear, first person, and rich academic language:

**Original Text:**

[transcribed from the spoken audio]

Throughout my trials and my reflections I have also come to realize that a few additional tools that might be developed in the future might not help again for increasing the efficiency of writing but also increasing transparency and making us more aware about our responsibility, especially on the ecological sustainability as a result of the tool and also make it make our writing less prone to dissemination of fake information. Currently there are talks and attempts to create watermark tools that would allow people to easily understand if the text is generated through tools, AI tools like GPT. Although of course this would be quite useful, especially in contexts such as classroom assignments, which also aim to teach the content to students, it may not be extremely meaningful for other cases, especially for academic writing. I don't think that our aim or goal should be banning the utilization of tools like GPT in academic writing and a binary solution such as a watermark, which suggests only if GPT is used or not, wouldn't be an ideal solution. Instead of that, we would need, for example, text editors that are supported by GPT and can document very clearly the prompts, the outcome of the prompts, and make clear, visible and easily understandable visualizations of the comparisons of these three versions of the text, which are the prompt, the result and the edited version. Additionally, I think I would find extremely useful a tool that will inform me about a carbon footprint of my writing piece. Of course, if this can be also be compared, can be done in comparison with my possible carbon footprint If I didn't use CHET-GPT, it would be more comparable and measurable. And in that way, I think scientists would then may find the responsibility in themselves to use it in the most beneficial way to the environment. Another tool set that would be useful would be about preventing the dissemination of faulty information, especially integrating measures that would detect authoritative tone in the written text and understand if this would refer to a fact would be useful and might make also the utilization methods such as research body more valuable in academic writing. One way that one thing that would tools should provide is of course reliable facts, but also the scraping and suggesting real references along with creating correct references just based on the names of the papers or the links, something that the current GPT is not capable of doing. I mean now if you give a link to GPT and want to create a reference out of it, it would list some of the information as correct, for example maybe author names, but would still improvising title or the URL links. So these kind of things can be prevented with the integration of these kind of tools.

**Result**

Throughout my trials and reflections I have come to the realization that further tools developed in the future may not only help increase the efficiency of writing, but also increase transparency and make us more aware of our responsibility – particularly with regard to ecological sustainability – and make our writing less prone to the dissemination of fake information. Currently, there are talks and attempts to create watermark tools that would enable people to easily identify if the text is generated through AI tools such as GPT. Although this would be useful in certain contexts, including classroom assignments that focus on teaching content to students, its utility may not be as significant in other scenarios, particularly for academic writing. Rather than banning the utilization of tools like GPT in academic writing, a binary solution such as a watermark which only indicates whether GPT is used or not may not be the best approach; instead, we might need text editors that are supported by GPT and can clearly document prompts, the outcomes of those prompts, and make clear, visible, and easily understandable visualizations of the comparisons between the prompt, the result, and the edited version. Additionally, I would find it extremely useful to have a tool that informs me of the carbon footprint of my writing piece and allows me to compare it to my potential carbon footprint if I didn't use GPT. Through such a comparison, scientists would be more aware of their responsibility in using GPT in the most beneficial way to the environment. Another toolset which would be beneficial is one that prevents the dissemination of faulty information, integrating measures which detect authoritative tone in the written text and comprehend whether it refers to a fact. Such tools should provide reliable facts, but should also scrape and suggest real references, and create correct references based on the names of the papers or the links; something that current GPTs are not capable of doing.

**Final**                                                                                                                   **Edit:**

> Throughout my trials and reflections, I have come to recognize that further tools developed in the future may not only help increase the efficiency of writing, but also increase transparency and make us more aware of our responsibility — particularly with regard to ecological sustainability — and make our writing less prone to the dissemination of fake information. Currently, there are attempts to create watermark tools (Stokel-Walker & Noorden, 2023) that would enable people to easily identify if the text is generated through LLM models such as GPT-3.5. Although this would be useful in certain contexts, including classroom assignments that focus on teaching content to students, its utility may not be as significant in other scenarios, particularly for academic writing. A binary solution such as a watermark which only indicates whether GPT is used or not may not be the best approach; instead, we might need text editors that are supported by GPT and can clearly document prompts, the outcomes of those prompts, and make clear, visible, and easily understandable visualizations of the comparisons between the prompt, the result, and the edited version (as I have manually done in this piece.)
>
> Additionally, a tool that gives information of the carbon footprint of the writing piece and compare it to potential carbon footprint if the GPT was not used might be useful. Through such a comparison, scientists would be more aware of their responsibility in using GPT in a mindful way to the environment.
>
> Another toolset which would be beneficial is one that prevents the dissemination of faulty information, integrating measures which detect authoritative tone in the written text and comprehend whether it refers to a fact. Such tools should provide reliable facts, scrape and suggest real references, and create correct references based on the names of the papers or the links; something that current GPTs are not capable of doing.

**Visual Comparisons of Additional tools that can help with GPT based academic writing / Whole Section:**

Red: Removed | Green: Added

**Original to Result:**

> 1    Throughout my trials and ~~my~~ reflections ~~I have~~ **also** come to ~~the~~ realiz~~ation~~**e** that ~~a few additional tools that might be~~ **further tools** developed in the future m~~ight~~**ay** not ~~help again for~~ **only help** increas~~ing~~**e** the efficiency of writing, but also increas~~ing~~**e** transparency and mak~~ing~~**e** us more aware ~~about~~**of** our responsibility, ~~especially on the ecological sustainability as a result of the tool and also make it~~ - **particularly with regard to ecological sustainability - and** make our writing less prone to **the** dissemination of fake information. Currently, there are ~~talks and~~ attempts to create watermark tools that would ~~allow~~**enable** people to easily ~~understand~~**identify** if the text is generated through ~~tools,~~ **AI tools** ~~like~~**such as** GPT. Although ~~of course~~ this would be ~~quite useful, especially in contexts such as~~ **useful in certain contexts, including** classroom assignments, ~~which also aim to~~ **that focus on** teach~~the~~**ing** content to students, it~~s~~ utility may not be ~~extremely meaningful for other cases, especially for academic writing. I don't think that our aim or goal should be~~ **as significant in other scenarios, particularly for academic writing. Rather than** banning the utilization of tools like GPT in academic writing ~~and~~, **a** binary solution such as a watermark, which ~~suggests only if GPT is used or not, wouldn't be an ideal solution. Instead of that, we would need, for example,~~ **only indicates whether GPT is used or not may not be the best approach; instead, we might need** text editors that are supported by GPT and can clearly document ~~very clearly the~~ prompts, the outcome~~s~~ of th~~ose~~ prompts, and make clear, visible, and easily understandable visualizations of the comparisons ~~of these three versions of the text, which are~~between the prompt, the result, and the edited version. Additionally, ~~I think~~ I would find ~~it extremely useful to have~~ a tool that ~~will~~ inform~~s~~ me about ~~a~~of the carbon footprint of my writing piece. ~~Of course, if this can be also be compared, can be done in comparison with my possible and~~ allows me to compare it to my potential carbon footprint ~~i~~f I didn't use ~~CHET-~~GPT~~, it would be more comparable and measurable. And in that way, I think~~GPT. Through such a comparison, scientists would ~~then may find~~be more aware of their responsibility in ~~themselves to use it~~using GPT in the most beneficial way to the environment. Another tool~~set~~ ~~that~~which would be ~~useful would be abou~~beneficial is one that prevent~~ing~~s the dissemination of faulty information, ~~especially~~ integrating measures ~~that would~~which detect authoritative tone in the written text ~~and understand if this would refer to a fact would be useful and might make also the utilization methods such as research body more valuable in academic writing. One way that one thing that would~~comprehend whether it refers to a fact. Such tools should provide ~~is of course~~ reliable facts, but should also ~~the~~ scrap~~ing~~e and suggest~~ing~~ real references ~~along with~~ ~~nd~~ creat~~ing~~e correct references ~~just~~ based on the names of the papers or the links; something that ~~the~~ current GPT~~is not capable of doing. I mean now if you give a link to GPT and want to create a reference out of it, it would list some of the information as correct, for example maybe author names, but would still improvising title or the URL links. So these kind of things can be prevented with the integration of these kind of tool~~ss are not capable of doing.

**Difference (%): 59.33**

**Result to Final Edit:**

1 Throughout my trials and reflections, I have come to ~~the realization~~recognize that further tools developed in the future may not only help increase the efficiency of writing, but also increase transparency and make us more aware of our responsibility - particularly with regard to ecological sustainability - and make our writing less prone to the dissemination of fake information. Currently, there are ~~talks and~~ attempts to create watermark tools (Stokel-Walker & Noorden, 2023) that would enable people to easily identify if the text is generated through ~~AI too~~LLM model~~l~~s such as GPT-3.5. Although this would be useful in certain contexts, including classroom assignments that focus on teaching content to students, its utility may not be as significant in other scenarios, particularly for academic writing. ~~Rather than banning the utilization of tools like GPT in academic writing, a~~A binary solution such as a watermark which only indicates whether GPT is used or not may not be the best approach; instead, we might need text editors that are supported by GPT and can clearly document prompts, the outcomes of those prompts, and make clear, visible, and easily understandable visualizations of the comparisons between the prompt, the result, and the edited version. ~~Additionally, I would find it extremely useful to have (as I have manually done in this piece.)~~

2 Additionally, a tool that ~~gives~~ inform~~s me~~ation of the carbon footprint of ~~my~~the writing piece and ~~allows me to~~ compare it to ~~my~~ potential carbon footprint if ~~I didn't use GPT~~the GPT was not used might be useful. Through such a comparison, scientists would be more aware of their responsibility in using GPT in ~~the most beneficia~~a mindful way to the environment.

3 Another toolset which would be beneficial is one that prevents the dissemination of faulty information, integrating measures which detect authoritative tone in the written text and comprehend whether it refers to a fact. Such tools should provide reliable facts, ~~but should also~~ scrape and suggest real references, and create correct references based on the names of the papers or the links; something that current GPTs are not capable of doing.

**Difference (%): 18.63**

<div style="text-align: center;">**- END -**</div>

# -NEW PARAGRAPH/SECTION-

**Section:** Limitations
**Paragraph:** Whole Section
**Method:** Talk Textualizer

**Temperature Setting:** 0.7

**Prompt:**

> Turn the below text into academic writing, clear, first person, and rich academic language:

**Original Text:**

[transcribed from the spoken audio]

> This is a perspective piece and this piece where I try to reflect my personal experience on creating a full academic document by heavily relying on GPT-3. Of course all my opinions and refkectşıbs have been affected by my perspective and positionality as a design researcher. Thus, although I think the ways of using GPT-3 in academic writing I framed in this paper can be useful for academics from many disciplines, my reflections on those may not be relevant to everyone and generalizable to all. In my opinion, people who have been in the fields where the applied part of the science is heavily occupying the portion of the work and at the same time the expression of ideas are almost equally important, my reflections might be more relevant to those people who works on those fields. Still, this piece should be taken into consideration knowing that it has been a short-term trial since one of the reasons of this trial was to understand the efficacy and it has been based on personal and subjective experience of one researcher. Also, the content of the writing in this piece was relatively practical and did not require much philosophical scrunity. The utilization of GPT-3 may not be as useful in fields where complex thoughts need to be expressed with very precise or abstract language.

**Result**

> Drawing upon my own experience, I sought to reflect on the use of GPT-3 in the creation of an academic document. It is important to note that my reflections have been shaped by my perspective and positionality as a design researcher. Thus, while I believe that the ways of using GPT-3 in academic writing as discussed in this paper can be beneficial for scholars from a variety of disciplines, my musings may not be relevant to all and should not be considered as generalizable across the board. In my opinion, my reflections may be of greater relevance to those working in fields where applied science plays an important role, yet expression of ideas is equally valued. It must be noted that this trial was limited in duration, as one of its primary aims was to assess efficacy. Moreover, the content of the writing in this paper was relatively practical and did not require much philosophical scrutiny. It is also likely that GPT-3 may not be as useful in fields where complex thoughts need to be expressed using precise or abstract language.

**Final Edit:**

Drawing upon my own experience, I sought to reflect on the use of GPT-3.5 in the creation of an academic article in this perspective piece. It is important to note that my reflections have been shaped by my perspective and positionality as a design researcher. Thus, while I believe that the ways of using of GPT-3.5 in academic writing as discussed in this paper can be beneficial for scholars from a variety of disciplines, my reflections and opinions may not be relevant to all and should not be considered as generalizable across the board. In my opinion, my reflections may be of greater relevance to those working in fields where applied science plays an important role, yet expression of ideas is equally valued. Also, it must be noted that this trial was limited in duration, as one of its primary aims was to assess efficacy, and longer engagement with the tool might allow the generation of new ideas, practices and opinions. Moreover, the content of the writing in this paper was relatively practical and did not require much philosophical scrutiny. Thus, GPT-3.5 may not be as useful in fields where complex thoughts need to be expressed using precise or abstract language. Similar experiments might be conducted to understand and demonstrate its efficacy across disciplines.

**Visual Comparisons of Limitations / Whole Section:**

`Red: Removed` | `Green: Added`

**Original to Result:**

1. This is a perspective piece and this piece where I try to reflect my personal experience on creating a full academic document by heavily relying on GPT-3. Of course all my opinions andDrawing upon my own experience, I sought to reflect on the use of GPT-3 in the creation of an academic document. It is important to note that my refklectsiions have been affectshaped by my perspective and positionality as a design researcher. Thus, although I thinkwhile I believe that the ways of using GPT-3 in academic writing I framas discussed in this paper can be useful for academics from many disciplines, my reflections on those may not be relevant to everyone and generalizable to all. In my opinion, people who have been in thebeneficial for scholars from a variety of disciplines, my musings may not be relevant to all and should not be considered as generalizable across the board. In my opinion, my reflections may be of greater relevance to those working in fields where the applied part of the science is heavily occupying the portion of the work and at the same time the expression of ideas are almost equally important, my reflections might be more relevant to those people who works on those fields. Still, this piece should be taken into consideration knowing that it has been a short-term trial since one of the reasons of this trial was to understand the efficacy and it has been based on personal and subjective experience of one researcher. Alsoscience plays an important role, yet expression of ideas is equally valued. It must be noted that this trial was limited in duration, as one of its primary aims was to assess efficacy. Moreover, the content of the writing in this pieceaper was relatively practical and did not require much philosophical scrunity. The utilization oftiny. It is also likely that GPT-3 may not be as useful in fields where complex thoughts need to be expressed with veryusing precise or abstract language.

**Difference (%): 79.52**

**Result to Final Edit:**

1. Drawing upon my own experience, I sought to reflect on the use of GPT-3.5 in the creation of an academic documentarticle in this perspective piece. It is important to note that my reflections have been shaped by my perspective and positionality as a design researcher. Thus, while I believe that the ways of using of GPT-3.5 in academic writing as discussed in this paper can be beneficial for scholars from a variety of disciplines, my musingreflections and opinions may not be relevant to all and should not be considered as generalizable across the board. In my opinion, my reflections may be of greater relevance to those working in fields where applied science plays an important role, yet expression of ideas is equally valued. IAlso, it must be noted that this trial was limited in duration, as one of its primary aims was to assess efficacy, and longer engagement with the tool might allow the generation of new ideas, practices and opinions. Moreover, the content of the writing in this paper was relatively practical and did not require much philosophical scrutiny. It is also likely thatThus, GPT-3.5 may not be as useful in fields where complex thoughts need to be expressed using precise or abstract language. Similar experiments might be conducted to understand and demonstrate its efficacy across disciplines.

**Difference (%): 23.94**

**- END -**

# -NEW PARAGRAPH/SECTION-

**Section:** Conclusion
**Paragraph:** Whole Section
**Method:** Talk Textualizer

**Temperature Setting:** 0.7

**Prompt:**

> Turn the below text into academic writing, clear, first person, and rich academic language:

**Original Text:**

[transcribed from the spoken audio]

> I have written this piece heavily relying on GPT-3 and in doing so I tried to frame the ways I had used to reflect in which parts and what ways these methods were useful. I expressed my reflections on the efficacy of the tool both in my writing and thought process, its time efficiency, and I also mentioned some of the points regarding ethical use of GPT-3 and some methods and some basic methods for providing transparency in academic writing. I hope the ways I framed in this paper would be useful to researchers who are considering using GPT-3 in their academic writings and would prompt both practices and discussions on methods of using, using it in a transparent way, along with inspiring tools that might make it easier to implement more responsible ways of using it in academic writing. Overall I hope this piece would contribute to do debate among incorporation of AI supported natural language processing tools in academic writing both by increasing understanding on its efficacy and also responsible usage.

**Result**

> In this paper, I have reflected on my experience of using GPT-3 as a tool for academic writing and discussed some basic methods for providing transparency when using it. I have shared my reflections on the efficacy of the tool, its time efficiency, and discussed the ethical considerations of using GPT-3 for academic writing. I hope that my reflections will be of benefit to researchers who are considering using GPT-3 and will encourage the development of more responsible practices and methods for using AI-supported natural language processing tools in academia. I believe that this work will contribute to the debate surrounding the incorporation of technology into our scholarly work, by increasing understanding of its efficacy and the importance of responsible usage.

```
Final                                                                    Edit:
```

In this paper, I have reflected on my experience of using GPT-3.5 as a tool for academic writing and discussed some basic methods for providing transparency when using it. I have shared my reflections on the efficacy of the tool and discussed the ethical considerations of using GPT-3.5 for academic writing around the issues raised by previous studies I hope that my reflections will be of benefit to researchers who are considering using GPT-3.5 and will inspire the development of tools for more responsible practices and methods for using AI-supported natural language processing in academia. I believe that this work will contribute to the debate surrounding the incorporation of this technology into our scholarly work, by increasing understanding of its efficacy and possible ways of responsible usage.

## Visual Comparisons of Conclusion / Whole Section:

Red: Removed | Green: Added

**Original to Result:**

1 | I have written this piece/paper, I heavily relying on GPT-3 and in doing so I tried to frame the ways I had used to reflect in which parts and what ways these methods were useful. I expressed my reflections on the efficacy of the tool both in my writing and thought process, its time efficiency, and I also mentioned some of the points regarding ethical use of GPT-3 and some methods and some basic methods for providing transparency in academic writing. I hope the ways I framed in this paper would be useful to researchers who are considering using GPT-3 in their academic writings and would prompt both practices and discussions on/e reflected on my experience of using GPT-3 as a tool for academic writing and discussed some basic methods for providing transparency when using it. I have shared my reflections on the efficacy of the tool, its time efficiency, and discussed the ethical considerations of using GPT-3 for academic writing. I hope that my reflections will be of benefit to researchers who are considering using GPT-3 and will encourage the development of more responsible practices and methods of/or using, using it in a transparent way, along with inspiring tools that might make it easier to implement more responsible ways of using it in academic writing. Overall I hope this piece would AI-supported natural language processing tools in academia. I believe that this work will contribute to do/the debate among/surrounding the incorporation of AI supported natural language processing tools in academic writing both/technology into our scholarly work, by increasing understanding o/nf its efficacy and also/the importance of responsible usage.

**Difference (%): 91.74**

**Result to Final Edit:**

1. In this paper, I have reflected on my experience of using GPT-3.5 as a tool for academic writing and discussed some basic methods for providing transparency when using it. I have shared my reflections on the efficacy of the tool, ~~its time efficiency,~~ and discussed the ethical considerations of using GPT-3.5 for academic writing, around the issues raised by previous studies. I hope that my reflections will be of benefit to researchers who are considering using GPT-3.5 and will ~~encourag~~ inspire the development of tools for more responsible practices and methods for using AI-supported natural language processing tools in academia. I believe that this work will contribute to the debate surrounding the incorporation of this technology into our scholarly work, by increasing understanding of its efficacy and ~~the importance~~ possible ways of responsible usage.

`Difference (%): 15.81`

**-END-**

# -NEW PARAGRAPH/SECTION-

**Section:** Conclusion
**Paragraph:** Last three sentences of 1st Para
**Method:** Polisher

**Temperature Setting:** 0.7

**Prompt:**

> Copy-edit the following text, academic style, first person, clear, rich and flowing language:

**Original Text:**

[transcribed from the spoken audio]

> I have framed five ways of using GPT-3.5 for academic writing; Chunk Stylist, Research Buddy, Talk Textualizer, Bullet-to-Paragraph and Polisher. Among those, I have used Chunk Stylist and Talk Textualizer predominantly, since they significantly increased by effectiveness while writing, while still retaining my tone and ideas. I think that other methods should be used with more care since they induce more spontaneity and randomness and might consolidate the concerns of plagiarism, distribution of fake information or reinforce problematic issues such as hate speech if not used with care. I also documented all prompts, outcomes and final edits in a separate supplementary document, which can serve as an example of using GPT-3.5 in a transparent way.

**Result**

> I have identified five ways of using GPT-3.5 for academic writing: Chunk Stylist, Research Buddy, Talk Textualizer, Bullet-to-Paragraph and Polisher. Of these, I have predominantly used Chunk Stylist and Talk Textualizer to great effect, enabling me to write more effectively while still retaining my own tone and ideas. Caution should be exercised with the other methods, however, as they introduce more spontaneity and randomness, which could result in plagiarism, the dissemination of false information, or even the exacerbation of hateful rhetoric if not used responsibly. To ensure fidelity and transparency, I documented all prompts, outcomes, and final edits in a separate supplementary document.

**Final** **Edit:**

> I have identified five ways of using GPT-3.5 for academic writing: Chunk Stylist, Research Buddy, Talk Textualizer, Bullet-to-Paragraph and Polisher. Of these, I have predominantly used Chunk Stylist and Talk Textualizer to great effect, enabling me to write more effectively while still retaining my own tone and ideas. Caution should be exercised with the other methods, however, as they introduce more spontaneity and randomness, which could result in plagiarism, the dissemination of false information, or even the exacerbation of hateful rhetoric if not used responsibly. To ensure fidelity and transparency, I documented all prompts, outcomes, and final edits in a separate supplementary document, which can serve as an example of using GPT-3.5 in a transparent way.

**Visual Comparisons of Conclusion / Last three sentences of 1st Para:**

Red: Removed | Green: Added

**Original to Result:**

1 | have ~~fram~~identified five ways of using GPT-3.5 for academic writing~~;~~: Chunk Stylist, Research Buddy, Talk Textualizer, Bullet-to-Paragraph and Polisher. ~~Among~~Of th~~o~~ese, I have predominantly used Chunk Stylist and Talk Textualizer ~~predominantly, since they significantly increased by effectiveness while writing,~~to great effect, enabling me to write more effectively while still retaining my own tone and ideas. ~~I think that other methods~~Caution should be ~~u~~exercised with ~~more care since~~the other methods, however, as they introduce more spontaneity and randomness ~~and might consolidate the concerns of plagiarism, distribu~~, which could result in plagiarism, the dissemination of fa~~k~~lse information, ~~or reinforce problematic issues such as hate speech if not used with care. I also~~or even the exacerbation of hateful rhetoric if not used responsibly. To ensure fidelity and transparency, I documented all prompts, outcomes, and final edits in a separate supplementary document~~, which can serve as an example of using GPT-3.5 in a transparent way~~.

**Difference (%): 60.25**

**Result to Final Edit:**

1 | I have identified five ways of using GPT-3.5 for academic writing: Chunk Stylist, Research Buddy, Talk Textualizer, Bullet-to-Paragraph and Polisher. Of these, I have predominantly used Chunk Stylist and Talk Textualizer to great effect, enabling me to write more effectively while still retaining my own tone and ideas. Caution should be exercised with the other methods, however, as they introduce more spontaneity and randomness, which could result in plagiarism, the dissemination of false information, or even the exacerbation of hateful rhetoric if not used responsibly. To ensure fidelity and transparency, I documented all prompts, outcomes, and final edits in a separate supplementary document, which can serve as an example of using GPT-3.5 in a transparent way.

**Difference (%): 8.94**

**-END-**

| SECTION | PARA | METHOD 1 | TEMP | OR | RE |
|---|---|---|---|---|---|
| Abstract | Whole | Polisher | 0.9 | 66.97 | 20.36 |
| Introduction | Whole | Chunk Stylist | 0.9 | 81.77 | 33.66 |
| Background | 1 | Research Buddy | 0.7 | - | 42.01 |
| Background | Rest | Chunk Stylist / Bullet-to-Para | 0.9 | 93.33 | 71.44 |
| Procedure followed | 1 | Polisher | 0.9 | 56 | 11.9 |
| Procedure followed | 2 | Bullet-to-Paragraph | 0.9 | 68.7 | - |
| Procedure followed | 2 | Polisher | 0.9 | 24.23 | 65.06 |
| Ways of using gpt-3.5 for academic writing | 1 | Polisher | 0.7 | 71.78 | 27.6 |
| Chunk stylist | Whole | Chunk Stylist | 0.9 | 82.61 | 25.54 |
| Research buddy | Whole | Research Buddy | 0.9 | - | 39.02 |
| Bullet-to-paragraph | Whole | Bullet-to-Paragraph | 0.9 | 78.25 | 4.12 |
| Talk textualizer | Whole | Talk Textualizer | 0.9 | 84.26 | - |
| Talk textualizer | Whole | Polisher | 0.9 | 61.24 | 20.79 |
| Polisher | Whole | Polisher | 0.7 | 34.39 | 19.25 |
| Reflections on using gpt-3.5 for writing articles | 1 | Talk Textualizer | 0.7 | 98.54 | 30.68 |
| Reflections on using gpt-3.5 for writing articles | 1 | Polisher | 0.7 | 54.15 | 0.24 |
| Reflections on using gpt-3.5 for writing articles | 2 | Talk Textualizer | 0.9 | 89.62 | 53.93 |
| Reflections on using gpt-3.5 for writing articles | 2 | Talk Textualizer | 0.9 | 44.76 | 23.07 |
| Reflections on using gpt-3.5 for writing articles | 3 | Talk Textualizer | 0.9 | 85.65 | 46.52 |
| Reflections on using gpt-3.5 for writing articles | 4 | Talk Textualizer | 0.7 | 67.78 | 3.3 |
| Ways of using methods | Whole | Talk Textualizer | 0.2 | 85.64 | 38.64 |
| Transparency, agency and biases | Whole | Talk Textualizer | 0.7 | 63.58 | 47.33 |
| Ecological sustainability | Whole | Talk Textualizer | 0.7 | 72.69 | 36.65 |
| Additional tools that can help with gpt | Whole | Talk Textualizer | 0.7 | 59.33 | 18.63 |
| Limitations | Whole | Talk Textualizer | 0.7 | 79.52 | 23.94 |
| Conclusion | Whole | Talk Textualizer | 0.7 | 91.74 | 15.81 |
| Conclusion | 1 | Polisher | 0.7 | 60.25 | 8.94 |
| | | | **Average** | **70.2712** | **29.1372** |